\documentclass{article}
\usepackage[utf8]{inputenc}
\usepackage{authblk}
\usepackage{setspace}
\usepackage[margin=1.25in]{geometry}
\usepackage{graphicx}
\graphicspath{ {./figures/} }
\usepackage{subcaption}
\usepackage{amsmath}
\usepackage{lineno}

\usepackage{url}
\usepackage[hidelinks]{hyperref}
\usepackage{amsfonts}
\usepackage{amssymb}
\usepackage{algorithm}
\usepackage[noend]{algpseudocode}
\usepackage{multirow}
\usepackage{tikz}
\usetikzlibrary{positioning}
\usetikzlibrary{arrows}
\mathchardef\mhyphen="2D

\newtheorem{proposition}{Proposition}
\newtheorem{example}{Example}
\newtheorem{theorem}{Theorem}
\newtheorem{corollary}{Corollary}
\newtheorem{definition}{Definition}

\usepackage[style=ieee, 
citestyle=numeric-comp,
sorting=none]{biblatex}
\title{Learning Probabilistic Temporal Safety Properties from Examples in Relational Domains}

\author[1]{Gavin Rens}
\author[1]{Wen-Chi Yang}
\author[2]{Jean-Fran\c{c}ois Raskin}
\author[1]{Luc De Raedt}

\affil[1]{KU Leuven University, Belgium}
\affil[2]{Universit\'e Libre de Bruxelles, Belgium}

\date{}

\begin{document}

\maketitle

\begin{abstract}
\noindent
We propose a framework for learning a fragment of probabilistic computation tree logic (pCTL) formulae from a set of states that are labeled as safe or unsafe. We work in a relational setting and combine ideas from relational Markov Decision Processes with pCTL model-checking. More specifically, we assume that there is an unknown relational pCTL target formula that is satisfied by only safe states, and has a horizon of maximum $k$ steps and a threshold probability $\alpha$. The task then consists of learning this unknown formula from states that are labeled as safe or unsafe by a domain expert. We apply principles of relational learning to induce a pCTL formula that is satisfied by all safe states and none of the unsafe ones. This formula can then be used as a safety specification for this domain, so that the system can avoid getting into dangerous situations in future. Following relational learning principles, we introduce a candidate formula generation process, as well as a method for deciding which candidate formula is a satisfactory specification for the given labeled states. The cases where the expert knows and does not know the system policy are treated, however, much of the learning process is the same for both cases. We evaluate our approach on a synthetic relational domain.

\end{abstract}


\section{Introduction}
Many recent publications report on methods for achieving safety in Markov Decision Processes (MDPs), where temporal logic (safety) specifications must be satisfied \cite{abeknt18,ccsm18,dfip19,hka19}.  
However, it is typically assumed that 1) the safety specification is given, and 2) that the states in the underlying MDP are unstructured. 
In this paper, we are interested in  1) {\em learning } the safety specification from examples,
and 2) working with relational MDPs.
More specifically, 
in our learning setting we assume that there is a domain expert who is presented with a set of system states $E$, a probability threshold $\alpha$ and a step-bound $k$ (number of action executions).
If the expert believes that the system, starting in $s\in E$ will perform actions that lead to a dangerous temporal situation within $k$ steps with probability at least $\alpha$, then she will label $s$ as dangerous, else, as safe. 


Now, given this set $E$ of labeled states, we want to learn a compact temporal logic formula summarizing the expert's advice. There are at least three reasons to infer a property (expressed as a temporal logic formula) from an expert's advice. Firstly, to obtain a concise, human-interpretable expression of some aspects of the domain \cite{lpb15,ng18,cm19}, secondly, to verify a system's control behavior (policy) w.r.t.\ a set of (safety) standards \cite{ng18,kjagb14} and thirdly, to use the (safety) property to devise strategies for the system or agent to avoid undesirable situations \cite{kjagb14,bvpyb16,vidbwdb17}.


Furthermore, we consider systems that can be modelled as relational MDPs (RMDPs). In other words, for the systems we consider, one can define the set of (relational) states and the stochastic effects (state transitions) for each system action. Although a reward model is often a component of an MDP, rewards do not play a part in this work. That is, we shall focus on system dynamics, not system utility.



Situations, including critical situations, often have a temporal component and a probabilistic component to them. To illustrate our setting, consider that
Lithium (used in most modern batteries) releases (flammable) hydrogen when coming in contact with water, or Rubidium, which has an even more immediate, explosive reaction when coming in contact with water.
A statistical analysis on data might show that if Lithium and water-based substance are not separated within $t$ minutes of discovering that they are in the same truck, that hydrogen is likely to be produced with certainty $c$. We might thus want to specify that containers holding Lithium and containers holding a water-based substance must be (put) in different trucks within 20 steps (minutes) with 80\% certainty.
Besides the chemical industry, most domains have potential dangerous situations, like industrial and service robots, warehouse packing, and vehicle driving.

We use a variant of computation tree logic (CTL) \cite{ce86} to represent temporal (including sequential) aspects of system dynamics. Furthermore, because we are interested in stochastic domains, we must be able to express the \textit{probability} of a sequence of states occurring. Hence, we shall use probabilistic CTL (pCTL) \cite{hj94}.
The safety property above could be expressed in pCTL as
\[
\mathsf{P}_{\geq 0.8}\mathsf{G}^{\leq 20}
[
\mathtt{lithium(X) \land waterBased(Z) \land in(X, T1) \land in(Z, T2)}
]
\]
where $\mathtt{X}$ is a container with Lithium in it, $\mathtt{Z}$ is a container with a water-based substance in it, $\mathtt{in(C,T)}$ means container $\mathtt{C}$ is in truck $\mathtt{T}$, and where variables with different names refer to different objects.
In this paper, we address the problem of inferring such a temporal logic formula that can be used to distinguish between safe and unsafe states.
A system or agent is always operating under a (possibly implicit) policy. We shall consider cases where the expert knows the underlying policy and where she does not know the policy. When the expert knows the policy, she can mark states as (un)safe \textit{with respect to} the policy she has in mind. When the expert does not know the policy, she will be more cautious, marking states as (un)safe if there is \textit{some} policy that leads to an (un)safe situation. If she knows the policy, a learnt pCTL property must be used with respect to that policy. Else, if she does not know the policy, the learnt property can be used for any/all policies.

Whether a learnt property is associated with a particular policy has benefits and drawbacks.
For instance, we might elicit the following property from an expert who has no particular policy in mind.
``With high probability, a driver must always (for the next 999 minutes) maintain a velocity such that the car can stop before reaching a red traffic light.'' This property can then be set as a guideline for all drivers. Now suppose that the expert is labeling states while assessing James's driving ability.
If we learn a property $\Psi$
(associated with James's policy $\pi_J$), then we can compare $\Psi$ with a set of policy-agnostic guidelines to decide whether James passes or fails.

Our contribution is \textit{the first method for inductively learning a relational probabilistic CTL formula from a set of positive and negative examples in an RMDP setting.} We call our algorithm \textit{Learn-pCTL}.
It combines techniques from relational learning, with RMDPs and pCTL model-checking. 

Regarding RMDPs, we build upon pCTL-REBEL  \cite{yrd21}, a recent  framework   for model-checking in relational MDPs. Given a pCTL formula, pCTL-REBEL computes the set of (relational) states in which the formula is satisfied. Given a candidate formula (candidate safety property), we employ pCTL-REBEL to determine whether all the states marked safe and none of the states marked dangerous satisfy the candidate. 

Regarding relational learning, we take inspiration mostly from the SeqLogMine algorithm \cite{ld04}. In particular, we adapt their refinement operator.

Next, we review the necessary theory and introduce some formalisms to facilitate reasoning with relational logic, and briefly review pCTL-REBEL.
In Section~\ref{sec:Learning-a-Relational-pCTL-Formula-from-Labeled-Examples}, we presented our approach, including the candidate formula generation process and how to check the consistency of a candidate w.r.t.\ the set of expert-labeled states.
Along the way, we also discuss three opportunities for pruning the space of candidates.
The algorithm and some theoretical results are also provided at the end of Section~\ref{sec:Learning-a-Relational-pCTL-Formula-from-Labeled-Examples}.
In Section~\ref{sec:Theoretical-Results}, we presents some theoretical results.
Section~\ref{sec:Experiments} validates our approach on a synthetic domain and Section~\ref{sec:policy-specific-algo}.
We end with a review of related work and then conclude with a summary of our research and a discussion about how it could be extended.

\section{Preliminaries}

\subsection{Relational Logic}
In this section, we present the necessary concepts of relational logic \cite{nd97,d08c}. 

An alphabet $\Sigma=\langle R,D \rangle$ is a finite set $R$ of relation symbols $r$, each with arity $m_r\geq 0$ and a possibly infinite set $D$ of constants, called the domain. An $m$-ary atom $r(t_1, ..., t_m)$ is a relation symbol $r$ followed by an $m$-tuple of terms $t_i$. A term is a variable $X$ or a constant $c$. A variable (resp.\ constant) is expressed by a string that starts with an upper (resp.\ lower) case letter. A conjunction $C$ is a set of atoms, and is implicitly assumed to be \textit{existentially quantified}.
Given an expression $E$, $vars(E)$ (resp.\ $consts(E)$, $terms(E)$) denotes the set of all variables (resp.\ constants, terms) in $E$. An expression is called \textit{ground} if it contains no variables. We shall call an expression \textit{sky} if it contains only variables. We shall make the unique name assumption, that states all constants are unequal, that is, $c_1 \neq c_2$ holds for different constants $c_1$ and $c_2$.

A substitution $\theta$ is a set of bindings $\{X_1 \gets t_1 , \ldots, X_n \gets t_n \}$ that assigns terms $t_i$ to variables $X_i$.
A conjunction $C$ with substitution $\theta$ applied to it (denoted $C\theta$) is obtained by simultaneously replacing all variables $X_i$ by their corresponding terms $t_i$.
As is done by \cite{yrd21},
we assume the Object Identity subsumption framework (OI-subsumption) of \cite{ffmb02}, which requires that any two terms $\mathtt{t_1,t_2}$ in an atom or conjunction are unequal.
For instance, under OI-subsumption the conjunction $\{\mathtt{{cl(b),on(X,Y)}}\}$ implicitly denotes the expression
$\{\mathtt{cl(b), on(X, Y), X \neq Y, X\neq b, Y\neq b}\}$.\footnote{The notation $\{a_1, a_2, \ldots, a_n\}$ is the set-theoretic form of the conjunction $a_1\land a_2\land \cdots\land a_n$. The two notations are used interchangably in this paper.}


A conjunction $C$ is OI-subsumed by conjunction $C'$, denoted $C \preceq_\theta C'$, iff $C'\theta \subseteq C$ under OI-subsumption.
And a conjunction $C$ is (generally) subsumed by conjunction $C'$, denoted $C \preceq C'$, if there \textit{exists} a substitution $\theta$ such that $C \preceq_\theta C'$.\footnote{\cite{yrd21} use notation $C \preceq_\theta C'$ to denote general subsumption (without specifying $\theta$).}
If $C'\theta$ is a proper subset of $C$, then $C'\theta$ and $C'$ are more general statements than $C$, that is, $C$ has more specific information that $C'\theta$ and $C'$.
For instance, $\{\mathtt{on(Y,X)}\}$ $\theta$-subsumes $\{\mathtt{on(a,X),cl(b)}\}$ for $\theta = \{\mathtt{Y\gets a}\}$ and for $\theta = \{\mathtt{Y\gets a, X\gets c}\}$ and even for $\theta = \{\mathtt{Y\gets a, X\gets Z}\}$.
Note that $\mathtt{Y}$ and $\mathtt{X}$ must be assigned different constant or variable, because under OI-subsumption, $\mathtt{Y}\neq\mathtt{X}$.
If there exists a substitution $\theta$ for which $C'\theta=C$, but $C'$ and $C$ are not identical, then $C'$ is also more general than $C$ ($C$ is more specific than $C'$). For instance, $\{\mathtt{on(Y,X), cl(b)}\}$ is more general than $\{\mathtt{on(a,X), cl(b)}\}$. 


The domain $D$ of an alphabet $\Sigma$ is the set of all constants in $\Sigma$. The Herbrand base of $\Sigma$, denoted as $HB^\Sigma$, is the set of all ground atoms that can be constructed from $\Sigma$. A Herbrand interpretation $I$ is a subset of $HB^\Sigma$ with intended meaning that an atom is in $I$ if and only if it is true. The set of all Herbrand interpretations of $\Sigma$ is denoted as $I^\Sigma$. When $D$ is infinite, $I^\Sigma$ is an infinite set.

In this work, we also consider expressions with some (or only) variables. Let $\mathit{Var}$ be a countable and infinite set of variables. The \textit{sky base} $SB^R$ is the set of all atoms that can be constructed with symbols in $R$ and only variables in $\mathit{Var}$. Hence, for $R = \{\mathtt{cl/1,on/2}\}$, $\mathtt{on(Y,X)}, \mathtt{on(Y,Y)}$, $\mathtt{on(X,X)}$, $\mathtt{on(X,Y)}$, $\mathtt{on(Z,Y)}$, $\mathtt{on(Z,Z)}$, $\ldots,$ $\mathtt{cl(X)}$, $\mathtt{cl(Y)}$, $\mathtt{cl(Z)}\ldots$ are in $SB^R$ for $X,Y,Z\ldots\in \mathit{Var}$. Let $\Theta$ be the set of all substitutions that can be constructed from $\mathit{Var}$ and $D$, including the empty substitution. Then we define $\mathit{AR} = \{A\theta\mid A\in SB^R \land \theta\in\Theta\}$, the set of all (partially ground) atomic relations. 
Note that $HB^\Sigma\subset \mathit{AR}$ and $SB^R\subset \mathit{AR}$.

\subsection{Relational MDPs}

The formalisms and notation in this section are aligned with those in \cite{yrd21}, who aligned their work with \cite{kvd04}. For a broader presentation about relational MDPs, the reader may refer to \cite{v08}.

A Markov decision process (MDP) is a tuple $\langle S,A,T,Rew\rangle$
with a set of states, a set of actions, a transition function, respectively, a reward function. In this work, the reward function does not play a role; we thus ignore it in the definition of MDPs.

A (non-rewarding) relational MDP (RMDP) is a pair $M=\langle \Sigma,\Delta\rangle$ where $\Sigma=\langle R,D \rangle $ is an alphabet containing a set of relations $R$ and a domain $D$, and $\Delta$ is a finite set of transition rules. $R$ can be divided into \textit{state relations} and \textit{action relations}. A state $s\in I^\Sigma$ is a Herbrand interpretation involving only state relations. An \textit{abstract state} $s'\subset \mathit{AR}$ is then a conjunction of atoms representing a set of Herbrand interpretations, $\{s\in I^\Sigma\mid  s\preceq s'\}$.
For instance, in a blocks world with relations $R = \{\mathtt{cl/1,on/2}\}$ and domain $D = \{\mathtt{bl_1, bl_2, bl_3}\}$, the abstract state $\{\mathtt{cl(X), cl(Z), on(X, Y)}\}$ represents a set of six physically possible ground states:
\begin{align*}
&	\{\mathtt{\{cl(bl_1), cl(bl_3), on(bl_1,bl_2)\},\{cl(bl_1), cl(bl_2), on(bl_1,bl_3)\}},\\
&	\mathtt{\{cl(bl_2), cl(bl_3), on(bl_2,bl_1)\},\{cl(bl_2), cl(bl_1), on(bl_2,bl_3)\}},\\
&	\mathtt{\{cl(bl_3), cl(bl_2), on(bl_3,bl_1)\},\{cl(bl_3), cl(bl_1), on(bl_3,bl_2)\}}\}.
\end{align*}

An abstract action is an atom for an action relation that does not belong to a Herbrand interpretation, that is, is not a state feature.
The following is an example by \cite{kvd04} of abstract transitions for the $\mathtt{move}$ action in the blocks world. It is specified by two rules $\delta_{move_1},\delta_{move_2}\in\Delta$. The abstract action $\mathtt{move(A, B, C)}$ succeeds in moving block $\mathtt{A}$ to block $\mathtt{B}$ from block $\mathtt{C}$ with probability $\alpha_1 = 0.9$ and fails with probability $\alpha_2 = 0.1$. 
When the action fails, the state stays the same.
\begin{align*}
& \delta_{move_1}: \mathtt{cl(A),cl(B),on(A,B) \stackrel{0.9:move(A, B, C)}{\longleftarrow}  cl(A),cl(B),on(A,C)}\\
& \delta_{move_2}: \mathtt{cl(A),cl(B),on(A,C) \stackrel{0.1:move(A, B, C)}{\longleftarrow}  cl(A),cl(B),on(A,C)}
\end{align*}

A RMDP generalizes a traditional MDP in two ways. First, a state in a MDP is represented by a constant. By allowing the use of conjunctions of ground atoms, a state in a RMDP is represented by a Herbrand interpretation. Second, by allowing the use of variables, a set of states in a RMDP can then be represented by an abstract state.

A RMDP $K=\langle \Sigma,\Delta\rangle$ defines an underlying ground MDP $\langle S,A,T\rangle$ where
$S$ is a set of Herbrand interpretations $I^\Sigma$ formed from state relations, $A$ is a set of ground actions induced from relational actions in $\Sigma$ and constants in $\Delta$, and $T$ is, similarly, a set of ground transitions from states in $I^\Sigma$ to states in $I^\Sigma$.
A stochastic, non-Markovian, stationary policy is a function $\pi: S^*\times A \to [0,1]$

\subsection{The $L^\mathit{FG}$ Fragment of Relational pCTL}

Probabilistic Computation Tree Logic (pCTL) is a temporal logic based on (non-probabilistic) CTL \cite{bk08b}.
Relational pCTL defined by \cite{yrd21}.
$L^\mathit{FG}$ is a non-nested, step-bounded, relational pCTL with temporal operators $\mathsf{F}$ (Finally) and $\mathsf{G}$ (Globally). 

$L^\mathit{FG}$ is defined as follows. Given an alphabet $\Sigma$ and set of variables $\mathit{Var}$ (inducing $\mathit{AR}$),  $L^\mathit{FG}$ formulae have the form
\[
\Psi = \mathsf{P}_{\geq \alpha}\Phi,
\]
where $\Phi$ is a \textit{path} formula and $\alpha\in[0,1]$ is a probability 
pCTL path formulae are formed according to the following grammar.
\[
\Phi ::= \mathsf{F}^{\leq k}\phi \mid \mathsf{G}^{\leq k}\phi
\]
where $\phi$ is a state formula defined by the grammar
\[
\phi ::= \mathit{ra} \mid \phi \land \phi
\]
where $\mathit{ra}\in \mathit{AR}$ and $k\in \mathbb{N}$. Note that any $\phi$ is an abstract state of an RMDP.

The informal semantics are
\begin{itemize}
\item $\mathsf{F}^{\leq k}\phi$: $\phi$ will be satisfied within $k$ steps.
\item $\mathsf{G}^{\leq k}\phi$: $\phi$ is always satisfied, for $k$ steps.
\end{itemize}

The formal semantics of $L^\mathit{FG}$ is the same as for the logic used with pCTL-REBEL \cite{yrd21}, given next.

We define the length of a formula $\Psi$ as the number of atoms in its state formula. For instance $\mathit{length}(\mathsf{P}_{\geq0.6}\mathsf{F}^{\leq4}[\mathtt{cl(a),on(b,c)}]) = 2$.

There is an issue when aiming to learn a property with an upper-bounded threshold probability. For instance, one might think that having target \\$\mathsf{P}_{< 0.1}\mathit{Op}^{\leq 9}\{\mathtt{on(X,Y),on(Y,Z)}\}$ makes sense, but because pCTL-REBEL seeks the policy that maximizes threshold probability, there are typically few or no satisfying states. These `upper-bound' formulae are thus excluded from our language.

\subsection{Semantics of pCTL-REBEL}
\label{sec:Semantics-of-pCTL-REBEL}

pCTL-REBEL \cite{yrd21} is a framework for performing model-checking on relational pCTL formulae.
The model checking task in pCTL-REBEL is to take a formula $\Psi$, a RMDP $K$ and then identify the set of abstract states in the associated MDP $M$ that satisfy $\Psi$, denoted as $Sat_M(\Psi)$.\footnote{To be precise, pCTL-REBEL can deal with RMDPs with infinite domains; it thus defines a \textit{b-bounded} MDP $M$ for a given RMDP. A b-bounded MDP $M^b = \langle S^b,A^b,T^b\rangle$ contains at most $b$ constants from $\Sigma$ (cf. \cite{yrd21}, Sect.\ 4 for details).} 
In pCTL-REBEL, $s$ satisfies $\Psi$ if there exists a policy and there exists a substitution such that $\Psi$ is true when starting in $s$.

The probability of any finite or infinite measurable set of paths $\Gamma$ (with the same starting state) through MDP $M$ under policy $\pi$ is denoted $P^\pi_M(\Gamma)$ (cf.~\cite{bk08b,yrd21} for definition).
Let $G$ be a set of \textit{goal states} in $S$ of MDP $M$.
Let $Path_{M,s}$ be all finite and infinite paths through $M$ starting in state $s$.
Then $Path_{M,s}(G)\subseteq Path_{M,s}$ is a set of paths that reach $G$ from $s$.
The maximum probability of reaching $G$ from $s$ in MDP $M$ for any policy $\pi$ in the policy space $\Pi$ of the MDP is
\begin{equation}
\label{eq:2.1}
P^{max}_{M} (Path_{M,s}(G)) \doteq \sup_{\pi\in\Pi} P^{\pi}_{M}(Path_{M,s}(G)).
\end{equation}

pCTL is based on the $\mathsf{X}$ (Next) and $\mathsf{U}$ (Until) operators. The $\mathsf{F}$ and $\mathsf{G}$ operators are then derived from these.
Formally, in pCTL-REBEL, given an alphabet, a Herbrand interpretation (state) $s$ and a relational pCTL formula $\Psi$, $s$ satisfies $\Psi$ if and only if there exists a grounding substitution $\theta$ for all free variables in $\Psi$ (under OI-subsumption) such that $s$ satisfies $\Psi$ under $\theta$, that is, $s\models \Psi \iff \exists\theta.s\models^\theta \Psi$.
The semantics of pCTL for pCTL-REBEL is then partially defined by
\begin{align}
s\models^\theta ra &\iff \mathit{ra}\theta \in s,\nonumber \\
s\models^\theta \phi \land \phi' &\iff s\models^\theta \phi \mbox{ and } s\models^\theta \phi',\nonumber \\
s\models^\theta \mathsf{P}_{\geq \alpha}\Phi &\iff P^{max}_M(\{\sigma\in Path_{M,s} \mid \sigma \models^\theta \Phi\})\geq \alpha,\label{eq:3.1}
\end{align}
where $\mathit{ra}$ is a relational atom in $\mathit{AR}$, and
\begin{align*}
\sigma\models^\theta \mathsf{X}\phi &\iff \sigma(2)\models^\theta\phi\\
\sigma\models^\theta \phi\mathsf{U}^{\leq k}\phi' &\iff \exists i \leq k. \sigma(i)\models^\theta\phi' \mbox{ and }\forall j<i. \sigma(j)\models^\theta\phi,
\end{align*}
where $\sigma(i)$ is the $i$-th state $s_i$ in path $\sigma$. 
Then $\mathsf{F}^{\leq k}\phi$ abbreviates $\top\mathsf{U}^{\leq k}\phi$ and $\mathsf{P}^{(\pi)}_{\geq \alpha}\mathsf{G}^{\leq k}\phi$ abbreviates $\mathsf{P}^{(\pi)}_{\leq 1-\alpha}\mathsf{F}^{\leq k}\lnot\phi$.

Recall that $Sat_M(\Psi) \doteq \{s\in I^\Sigma\mid s\models \Psi\}$.
In practice, pCTL-REBEL computes $Sat_K(\Psi)$ where
\begin{equation}
s\in Sat_M(\Psi) \iff \exists s_a \in Sat_K(\Psi): s \preceq s_a,
\label{eq:Sat}
\end{equation}
where $s\in I^\Sigma$ and $s_a\in 2^\mathit{AR}$.
To be precise, the version of pCTL-REBEL presented by \cite{yrd21} defines only the `policy-agnostic' operator $\mathsf{P}_{\geq \alpha}$. For this work, we extended pCTL-REBEL to also define the `policy-specific' operator $\mathsf{P}^\pi_{\geq \alpha}$. 
\begin{equation}
\label{eq:3.2}
    s\models^\theta \mathsf{P}^\pi_{\geq \alpha}\Phi \iff P^\pi_M(\{\sigma\in Path_{M,s} \mid \sigma \models^\theta \Phi\})\geq \alpha.
\end{equation}
Moreover, we write $Sat_M(\Psi)$ and $Sat_K(\Psi)$ when the policy is unknown (and $\mathsf{P}_{\geq \alpha}$ is used), and we write $Sat^\pi_M(\Psi)$ and $Sat^\pi_K(\Psi)$ when the policy is known (and $\mathsf{P}^\pi_{\geq \alpha}$ is used).

Although $\mathsf{P}_{\geq \alpha}\Phi$ is fully defined by now, simply to clarify its semantics, we offer an alternative characterization:
Let $P^{\pi,\theta}_{M,s}(\Phi)$ be the probability of $\Phi$ in MDP $M$ at concrete state $s$ with substitution $\theta$ under policy $\pi$.
Then from \eqref{eq:3.1} we get
\[
s\models \mathsf{P}_{\geq\alpha}\Phi \iff \exists\pi.\exists\theta.P^{\pi,\theta}_{M,s}(\Phi)\geq\alpha.
\]
and from \eqref{eq:3.2} we get
\[
s\models \mathsf{P}^\pi_{\geq\alpha}\Phi \iff \exists\theta.P^{\pi,\theta}_{M,s}(\Phi)\geq\alpha.
\]
For instance,
\begin{align*}
s\models \mathsf{P}_{\geq0.8}\mathsf{F}^{\leq 4} \mathtt{on(X,b)} &\iff \exists\pi.\exists\theta.P^{\pi,\theta}_{M,s}(\mathsf{F}^{\leq 4} \mathtt{on(X,b)})\geq0.8.
\end{align*}

The algorithm for computing which states satisfy a given formulae (for \textit{some} policy) is based on \textit{relational value iteration}. 
\cite{yrd21} employ a version of the RElational BELman (REBEL) update operator developed by \cite{kvd04}. The REBEL operator is used to define MDP value iteration for relational domains via logical regression.

\section{Learning a Relational pCTL Formula from Labeled Examples}
\label{sec:Learning-a-Relational-pCTL-Formula-from-Labeled-Examples}

We present the algorithm in several steps.
First, the learning problem is stated. Then we describe how a partial order can be defined over our target language, $L^\mathit{FG}$, inducing a subsumption lattice and how this can be used for pruning the search space.
Then, an optimal refinement operator is defined, which is used to generate candidates (i.e. search the space of solution properties), via four subsections.
Before the actual, high-level algorithm is presented, we discuss pruning by domain knowledge and by checking semantic equivalence.

\subsection{The Learning Problem}
\label{sec:Consistency-Checking}

When the policy is unknown, the problem we want to solve can be stated as

\noindent
\textbf{Given:}
\begin{itemize}
\item RMDP $K$
\item threshold $\alpha\in[0,1]$
\item step-bound $k\in\mathbb{N}$
\item $E = E^+\cup E^-$ s.t.\ $E^+\subset 2^\mathit{AR}$ of abstract states, and $E^-\subset 2^\mathit{AR}$ of abstract states
\end{itemize}
\textbf{Find:}
\begin{itemize}
\item $\Psi\in L^\mathit{FG}$, where $\Psi = \mathsf{P}_{\geq \alpha}\mathit{Op}^{\leq k}\phi$
\item such that $\forall s^+_a\in E^+. \forall s\in I^\Sigma. (s \preceq s^+_a \implies s \models \Psi)$
\item and $\forall s^-_a\in E^-. \forall s\in I^\Sigma. (s \preceq s^-_a \implies s \not\models \Psi)$
\item (in which case, $\Psi$ is said to be \textit{consistent} with $E$)
\item (where $\preceq$ is OI-subsumption)
\end{itemize}

When the system's policy is known/given, we write $s\models_\pi \Psi$ to highlight that the satisfaction of $\Psi$ by $s$ depends on $\pi$.
Hence, when the policy is known, we add ``policy $\pi$'' to the ``Given'' part and we change $\models$ and $\not\models$ in the ``Find'' part to $\models_\pi$, respectively, $\not\models_\pi$. 

Why do we choose $s\preceq s^+_a$ instead of $s^+_a\preceq s$, and $s\preceq s^-_a$ instead of $s^-_a\preceq s$ in the ``Find'' part? Consider the following example. Note that $\mathtt{on(a,b)}\preceq\mathtt{on(X,b)}\preceq\mathtt{on(X,Y)}$. Now suppose the expert advises that $\mathtt{on(X,b)}$ is safe.
Then it is fine to learn a property $\Psi$ such that $\mathtt{on(a,b)}\models\Psi$, but not such that $\mathtt{on(X,Y)}\models\Psi$.
Next suppose the expert advises that $\mathtt{on(X,b)}$ is dangerous. 
If we consider candidate $\Psi$ such that $\mathtt{on(a,b)}\models\Psi$, then we must reject $\Psi$. In other words, $\Psi$ must be rejected if there exists a state $s$ subsumed by $s^-_a$ such that $s\models\Psi$. That is, to accept $\Psi$, there may not exists a state $s$ subsumed by $s^-_a$ such that $s\models\Psi$. Or, for all states $s$ subsumed by $s^-_a$, $s\not\models\Psi$.

The problem statement above gives a formal/semantic description of what a solution looks like, but it does not give much advice on how to solve the problem in practice. Proposition~\ref{prp:consistency} provides a route to using pCTL-REBEL to solve the problem. It transforms the problem from reasoning about all states in $I^\Sigma$ to reasoning about the existence of states in $Sat_K(\Psi)$.

\begin{proposition}
\label{prp:consistency}
$\forall s^+_a\in E^+. \forall s\in I^\Sigma. (s \preceq s^+_a \implies s \models \Psi)$ iff $\forall s^+_a\in E^+, \exists s^+_a\in Sat_K(\Psi). s_a^+\preceq s_a$. And $\forall s^-_a\in E^-. \forall s\in I^\Sigma. (s \preceq s^-_a \implies s \not\models \Psi)$ iff $\forall s^-_a\in E^-, \nexists s_a\in Sat_K(\Psi). s^-_a\preceq s_a$.
\end{proposition}
\textbf{Proof:}
\begin{align*}
  & \forall s^+_a\in E^+. \forall s\in I^\Sigma. (s \preceq s^+_a \implies s \models \Psi)\\
   \iff &\forall s^+_a\in E^+. \forall s\in I^\Sigma. (s \preceq s^+_a \implies s \in Sat_M(\Psi))\\
   \iff &\forall s^+_a\in E^+. \forall s\in I^\Sigma. (s \preceq s^+_a \implies \exists s_a \in Sat_K(\Psi).s\preceq s_a) \;(\mbox{by Equivalence } \ref{eq:Sat})\\
   \iff &\forall s^+_a\in E^+. \exists s_a \in Sat_K(\Psi).s^+_a\preceq s_a\\
   & (\mbox{because: if }\forall s\in I^\Sigma.s \preceq s^+_a \implies s\preceq s_a,\mbox{ then }s^+_a\preceq s_a)
\end{align*}
\begin{align*}
  & \forall s^-_a\in E^-. \forall s\in I^\Sigma. (s \preceq s^-_a \implies s \not\models \Psi)\\
   \iff &\forall s^-_a\in E^-. \forall s\in I^\Sigma. (s \preceq s^-_a \implies s \not\in Sat_M(\Psi))\\
   \iff &\forall s^-_a\in E^-. \forall s\in I^\Sigma. (s \preceq s^-_a \implies \nexists s_a \in Sat_K(\Psi).s\preceq s_a) \;(\mbox{by Equivalence } \ref{eq:Sat})\\
   \iff &\forall s^-_a\in E^-. \nexists s_a \in Sat_K(\Psi).s^-_a\preceq s_a\\
   & (\mbox{because: if }\forall s\in I^\Sigma.s \preceq s^-_a \implies s\preceq s_a,\mbox{ then }s^-_a\preceq s_a)
\end{align*}
\hfill $\blacksquare$ \vspace{3mm} \noindent

To determine consistency of $\Psi$ with respect to $E$ (for unknown policy) in terms of pCTL-REBEL, we use the definition of $Sat_K(\Psi)$: The learning problem can thus also be stated equivalently as
\begin{corollary}
\label{crl:1}
~

\noindent
\textbf{{\em Find:}}
\begin{itemize}
\item $\Psi\in L^\mathit{FG}$
\item {\em such that} $\forall s^+_a\in E^+, \exists s_a\in Sat_K(\Psi). s^+_a\preceq s_a$
\item {\em and} $\forall s^-_a\in E^-, \nexists s_a\in Sat_K(\Psi). s^-_a\preceq s_a$
\end{itemize}
Similarly for when the policy is known.
\end{corollary}


\begin{example}
As a simple example, suppose
\[
E^+ = \{[\mathtt{cl(a),on(a,b)}],[\mathtt{cl(a),on(a,c)}]\} , E^- = \{[\mathtt{on(X,b),on(b,c)}]\} \mbox{ and } Sat_K(\Psi) = \{[\mathtt{cl(a),on(a,Y)}]\}.
\]
Then, first considering $E^+$, we see that
\[
[\mathtt{cl(a),on(a,b)}]\preceq[\mathtt{cl(a),on(a,Y)}] \mbox{ and } [\mathtt{cl(a)}, \mathtt{on(a,c)}] \preceq [\mathtt{cl(a),on(a,Y)}].
\]
Then considering $E^-$, we see that
\[
[\mathtt{on(X,b),on(b,c)}]\not\preceq[\mathtt{cl(a)}, \mathtt{on(a,Y)}],
\]
 that is, there is no example in $E^-$ such that it is subsumed by a state in $Sat_K(\Psi)$.
Hence, $\Psi$ is consistent with $E$ and should be accepted as one of the solutions or properties of interest.

As an example of when a candidate is not a solution, consider the case where $Sat_K(\Psi) = \{[\mathtt{on(X,Y)}]\}$.
First, considering $E^+$, we see that
\[
[\mathtt{cl(a),on(a,b)}]\preceq[\mathtt{on(X,Y)}] \mbox{ and } [\mathtt{cl(a),on(a,c)}] \preceq [\mathtt{on(X,Y)}].
\]
Second, considering $E^-$, we see that
\[
[\mathtt{on(X,b),on(b,c)}]\preceq[\mathtt{on(X,Y)}],
\]
that is, there exists an example in $E^-$ such that it is subsumed by a state in $Sat_K(\Psi)$.
Hence, $\Psi$ is not consistent with $E$ and should be rejected.
\end{example}

The task of finding all consistent solutions is solved by generating ever more specific candidate formulae and checking whether they are consistent with the set of examples $E$. 
Our approach is a general to specific one. We start with the most general `formula' (the empty formula) and generate slightly more specific formulae. For each formula generated, if not pruned (to be discussed later), it is refined to be even more specific, and so on, until some stopping criterion is reached.
This is the approach taken in concept-learning \cite{mitchell1982generalization}, frequent pattern mining and \cite{ah14} clausal discovery applications \cite{de1997clausal}.


\subsection{Partial Ordering}

To facilitate optimizing the process of searching for a formula consistent with $E$, we establish a partial order relation over $L^\mathit{FG}$. The idea is similar to \cite{kjagb14} and \cite{f11} and has links to the work of \cite{ld04}. Intuitively, we want to establish a subsumption lattice over formulae in our target language. This is required to design a principled subsumption-based pruning strategy.

\begin{definition}[Relation $\preceq_\mathit{FG}$]
For two formulae $\Psi, \Psi' \in L^\mathit{FG}$, $\Psi'\preceq_\mathit{FG}\Psi \iff \mathit{Mod}(\Psi')\subseteq \mathit{Mod}(\Psi)$, where $\mathit{Mod}(\Psi)\doteq\{s\in I^\Sigma\mid s\models\Psi\}$. If $\Psi'\preceq_\mathit{FG}\Psi $, then we say that $\Psi'$ is subsumed by $\Psi$, or that  $\Psi$ subsumes $\Psi'$.
$\Psi'$ is strictly subsumed by $\Psi$ (notation: $\Psi'\prec_\mathit{FG}\Psi $) iff $\Psi'\preceq_\mathit{FG}\Psi $ and $\Psi\not\preceq_\mathit{FG}\Psi' $.
\end{definition}

To link Corollary~\ref{crl:1} to relation $\preceq_\mathit{FG}$, we have the following proposition. It is also required for the soundness of pruning by subsumption, discussed next.
\begin{theorem}
\label{prp:Sat-subset-Sat}
$\Psi\preceq_\mathit{FG}\Psi'$ implies $Sat^{(\pi)}_K(\Psi)\subseteq Sat^{(\pi)}_K(\Psi')$, where $^{(\pi)}$ indicates that both known and unknown policy cases are considered.
\end{theorem}
\textbf{Proof:}
$\Psi\preceq_\mathit{FG}\Psi'$ is defined as $\mathit{Mod}(\Psi)\subseteq \mathit{Mod}(\Psi')$, which implies that
\begin{eqnarray}
\Psi\preceq_\mathit{FG}\Psi' &\iff& \forall s\in I^\Sigma, \mbox{ if } s\models\Psi, \mbox{ then } s\models\Psi'\\
&\iff& \mbox{if }s\in Sat^{(\pi)}_M(\Psi), \mbox{ then } s\in Sat^{(\pi)}_M(\Psi')\\
&\iff& \mbox{if }\exists s_a\in Sat^{(\pi)}_K(\Psi): s\preceq s_a,\nonumber\\
&& \mbox{then } \exists s'_a\in Sat^{(\pi)}_K(\Psi'): s\preceq s'_a  \qquad\mbox{(by Eq.~\ref{eq:Sat})}\label{ln:3}
\end{eqnarray}
Note that if $s\preceq s_a \implies s\preceq s'_a$, then $s_a\preceq s'_a$.
Therefore, line~\ref{ln:3} implies that if $\exists s_a\in Sat^{(\pi)}_K(\Psi)$, then $\exists s'_a\in Sat^{(\pi)}_K(\Psi'): s_a\preceq s'_a$.
Now note that if $s_a\in Sat^{(\pi)}_K(\Psi)$, then $\forall s_a''.s_a''\preceq s_a$, $s_a''\in Sat^{(\pi)}_K(\Psi)$.
And we know that $s_a''\preceq s_a\preceq s_a'$. Hence, $s_a''\preceq s_a'$.
Therefore, $s_a''\in Sat^{(\pi)}_K(\Psi')$. That is, if $s_a''\in Sat^{(\pi)}_K(\Psi)$, then $s_a''\in Sat^{(\pi)}_K(\Psi')$, which implies that $Sat^{(\pi)}_K(\Psi)\subseteq Sat^{(\pi)}_K(\Psi')$.
\hfill $\blacksquare$ \vspace{3mm}

\subsection{Pruning by Subsumption}
\label{sec:Pruning-by-Subsumption}

The next two propositions state the relationship between relation $\preceq_\mathit{FG}$ between formulae in $L^{FG}$ and relation $\preceq$ between state formulae.

\begin{proposition}
\label{prp:subsumption-for-inst-tree}
Given two formulae $\mathsf{P}_{\geq \alpha}\mathit{Op}^{\leq k}\phi'$ and $\mathsf{P}_{\geq \alpha}\mathit{Op}^{\leq k}\phi$ (where $\mathit{Op}$ is either $\mathsf{F}$ or $\mathsf{G}$), if $\phi'\preceq \phi$, then $\mathsf{P}_{\geq \alpha}\mathit{Op}^{\leq k}\phi'\preceq_\mathit{FG}\mathsf{P}_{\geq \alpha}\mathit{Op}^{\leq k}\phi$.
\end{proposition}
\textbf{Proof:}
\,

$\mathbf{F}$ \textbf{operator:} Let $\Psi' = \mathsf{P}_{\geq \alpha}\mathsf{F}^{\leq k}\phi'$ and $\Psi = \mathsf{P}_{\geq \alpha}\mathsf{F}^{\leq k}\phi$ such that $\phi'\preceq \phi$. Assume $s\models\phi'$. Then, by the assumption of the $\mathsf{F}$ operator, there exists an integer $i\leq k$ such that $\sigma_s(i)\models\phi'$; let $i$ be the smallest such index. By the definition of $\phi'\preceq \phi$, $\exists j\leq i.\sigma_s(j)\models\phi$. Hence, the probability that $s'\models\phi$ for some $s'$ in $\sigma_s$ within $k$ steps from $s$ is at least the probability that $s''\models\phi'$ for some $s''$ in $\sigma_s$. Therefore, if $s\models \mathsf{P}_{\geq \alpha}\mathsf{F}^{\leq k}\phi'$, then $s\models \mathsf{P}_{\geq \alpha}\mathsf{F}^{\leq k}\phi$, which implies $Mod(\Psi')\subseteq Mod(\Psi)$, which implies $\Psi'\preceq_\mathit{FG}\Psi$.

$\mathbf{G}$ \textbf{operator:} Let $\Psi' = \mathsf{P}_{\geq \alpha}\mathsf{G}^{\leq k}\phi'$ and $\Psi = \mathsf{P}_{\geq \alpha}\mathsf{G}^{\leq k}\phi$ such that $\phi'\preceq \phi$. Assume $s\models\Psi'$. Then, by the definition of the $\mathsf{G}$ operator, for all integers $i\leq k$, $\sigma_s(i)\models\phi'$. By the assumption of $\phi'\preceq \phi$, the probability that $s'\models\phi$ for some $s'$ in $\sigma_s$ for all $k$ steps from $s$ is at least the probability that $s''\models\phi'$ for some $s''$ in $\sigma_s$. Therefore, if $s\models \mathsf{P}_{\geq \alpha}\mathsf{G}^{\leq k}\phi'$, then $s\models \mathsf{P}_{\geq \alpha}\mathsf{G}^{\leq k}\phi$, which implies $Mod(\Psi')\subseteq Mod(\Psi)$, which implies $\Psi'\preceq_\mathit{FG}\Psi$.
\hfill $\blacksquare$ \vspace{3mm}

\begin{proposition}
\label{prp:F-subsumes-G}
$\mathsf{P}_{\geq p}\mathsf{G}^{\leq k}\phi \preceq_\mathit{FG} \mathsf{P}_{\geq p}\mathsf{F}^{\leq k}\phi$.
\end{proposition}
\textbf{Proof:}
Let $\Psi' = \mathsf{P}_{\geq \alpha}\mathsf{G}^{\leq k}\phi$ and $\Psi = \mathsf{P}_{\geq \alpha}\mathsf{F}^{\leq k}\phi$. Assume $s\models\Psi'$. By the definition of the $\mathsf{G}$ operator, the probability that [for all $s'$ such that $s'=\sigma_s(i)$ for $i=1,\ldots, k$, $s'\models\phi$] is greater or equal to $\alpha$. And therefore, by the definition of the $\mathsf{F}$ operator, for $i=1,\ldots, k$, $s'\models\Psi$. Thus, $s\models\Psi$. That is, if $s\models\Psi'$, then $s\models\Psi$, which implies $Mod(\Psi')\subseteq Mod(\Psi)$, which implies $\Psi'\preceq_\mathit{FG}\Psi$.
\hfill $\blacksquare$ \vspace{3mm}

Proposition~\ref{prp:Sat-subset-Sat} allows us to use pCTL-REBEL to prune the search space according to Propositions~\ref{prp:subsumption-for-inst-tree} and \ref{prp:F-subsumes-G}.

Let $\Psi$ be a candidate formula.
Employing Proposition~\ref{prp:subsumption-for-inst-tree}, we can implement the following pruning strategy. 
For any candidate formula $\Psi$, if
$\exists s^+_a\in E^+.\nexists s_a\in Sat_K(\Psi).s^+_a\preceq s_a$,
then for all formulae $\Phi'\prec_\mathit{FG}\Phi$,
$\nexists s'_a\in Sat_K(\Psi').s^+_a\preceq s'_a$. Hence, prune the search at $\Psi$.

For instance, we know that $\mathtt{[lithium(X),in(X,Y)]}$ subsumes $\mathtt{[lithium(X)}$, $\mathtt{in(X,c2)]}$. These may be used to generate candidates $\Psi^m$ $=$\\ $\mathsf{P}_{\geq 0.8}\mathsf{G}^{\leq 3}\mathtt{\{lithium(X), in(X,Y)\}}$ and $\Psi^n = \mathsf{P}_{\geq 0.8}\mathsf{G}^{\leq 3}\mathtt{\{lithium(X), in(X,c2)\}}$. But if $\exists s^+_a\in E^+.\nexists s_a\in Sat_K(\Psi^m).s^+_a\preceq s_a$,
then prune the search at $\Psi^m$ and do not generate $\Psi^n$ or any of its specializations. This is pruning due to instantiation.

Or, in a scenario involving a security robot and a suspicious package, suppose \[\mathsf{P}_{\geq 0.9}\mathsf{F}^{\leq 10}\mathtt{\{suspicious(X)\}}\] is inconsistent with $E^+$, that is, there is no state in $E^+$ from which the robot finds a suspicious package within ten steps with probability 0.9. Then there is no use in generating/checking $\mathsf{P}_{\geq 0.9}\mathsf{F}^{\leq 10}\mathtt{\{suspicious(X)\}}\land \mathtt{in(X,sa5)}]$, that is, additionally checking whether the package can be moved to safety area 5. This is pruning due to lengthening.

Employing Proposition~\ref{prp:F-subsumes-G}, we can implement the following pruning strategy.
Given a conjunction $\phi^n$, generate candidate $\Psi^\mathsf{F} = \mathsf{P}_{\geq \alpha}\mathsf{F}^{\leq k}\phi^n$ and check it for consistency before generating and checking candidate $\Psi^\mathsf{G} = \mathsf{P}_{\geq \alpha}\mathsf{G}^{\leq k}\phi^n$.
If $\exists s^+_a\in E^+.\nexists s_a\in Sat_K(\Psi^\mathsf{F}).s^+_a\preceq s_a$,
then $\nexists s'_a\in Sat_K(\Psi^\mathsf{G}).s^+_a\preceq s'_a$, hence, do not generate and check $\Psi^\mathsf{G}$.


Pruning based on Proposition~\ref{prp:subsumption-for-inst-tree} is much more important than pruning based on Proposition~\ref{prp:F-subsumes-G}. The former can avoid generating/checking candidates based on sub-trees (potentially thousands of candidates), whereas the latter avoids only one candidate generation/check.

\subsection{Optimal Refinement Operator}

To formalise the search process, we define a refinement operator (as is typical in ILP \cite{nd97,d08c}).
The notion of an \textit{optimal} refinement operator in relational learning is due to \cite{db93}.

We define a (specializing) refinement operator $\rho$ as follows.
Given poset $(L^{FG}$, $\preceq_{FG})$ and $\Psi,\Psi'\in L^{FG}$, $\rho(\Psi) \subseteq \{\Psi'\mid \Psi'\prec_\mathit{FG}\Psi\}$, where $\prec_\mathit{FG}$ implies a strict refinement.
With such an operator, we can employ a level-wise algorithm to generate and check formulae.

An $n$-step refinement is defined as
\begin{eqnarray*}
\rho^1(\Psi) &=& \rho(\Psi),\\
\rho^n(\Psi) &=& \{\Psi'\mid \exists \Psi''\in \rho^{n-1}(\Psi) \;\&\; \Psi' \in\rho(\Psi'')\}.
\end{eqnarray*}

For optimality, we further require that \cite{db93,ld04}:

\paragraph{Completeness.} Repeatedly applying the operator $\rho$ on the top formula $\top$ (the most general formula), it is possible to generate all other formulae. In other words, 
$\bigcup^\infty_{r=0}\rho^r(\top)$ $= \{\Psi'\mid \Psi'\preceq_{FG}\top\}$. This requirement guarantees that we will not miss any formulae that may be consistent with $E$, as long as we start from the top formula.
\paragraph{Single Path.} Given formula $\Psi$, there should exist exactly one sequence of formulae $\Psi_0 = \top, \Psi_1,...$ $,\Psi_n = \Psi$ such that $\Psi_{i+1} \in\rho(\Psi_i)$ for all $i$. This requirement helps ensuring that no candidate is generated more than once, that is, there are no duplicates. 

\bigskip
\noindent
For Learn-pCTL, $\rho(\{\mathsf{P}_{\geq \alpha}Op^{\leq k}\phi)$ is defined via four operations.
\begin{itemize}
    \item Lengthening (Len): Add one atom to $\phi$.
    \item Unification (Uni): Unify one variable with a variable or an already existing constant.
    \item Instantiation (Ins): Instantiate one variable with a new constant.
    \item Globalization (Glo): Change an Eventually-formula to a Globally-formula.
\end{itemize}
Except for Glo, all refinement operations are applied to conjunction/state-formula $\phi$. 

In Len, Uni and Ins, there is potential for generating duplicate formulae, which would violate the {\em single path} property. In the next subsections we discuss duplicate-avoidance strategies in general, for each of Len, Uni and Ins. We end this section with a formal definition of $\rho$ and an algorithm showing how it is used.

\subsection{Formula Lengthening}

Larger/longer formulae (abstract states) are generated deeper in the tree.
Let $ord(R) = (r_1/a_1, r_2/a_2$, $\ldots, r_m/a_m)$ be an ordering of the relations in $R$. Every child of the root is an atom $r_i(X_1,\ldots,X_{a_i})$ where the $X_i$ are variables in $Var$. Every node in the tree is a list of atoms representing their conjunction. Every variable in a node has a different name. Variables in different nodes may have the same name.

The children of node $n$ are formed by adding one atom to the list from $n$ according to $ord(R)$. To avoid syntactic variants by generating permutations of the same atoms, we use the following expansion rule: If $n$ have list $(q_1/a_1, \ldots, q_t/a_t)$, then $n$ is expanded with $\ell$ children such that child $n'_1$ has list $(q_1/a_1, \ldots, q_t/a_t,r_i/a_i)$ where $q_t=r_i$, child $n'_2$ has list $(q_1/a_1, \ldots, q_t/a_t,r_{i+1}/a_{i+1})$, ... child $n'_\ell$ has list $(q_1/a_1, \ldots, q_t/a_t,r_m/a_m)$. 

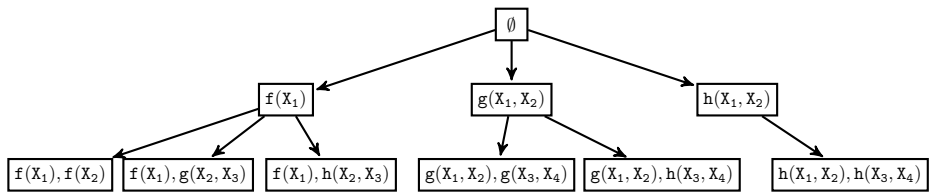
\begin{figure}
\centering
\begin{tikzpicture}[->,>=stealth',shorten >=1pt,auto,node distance=3cm,
  thick,main node/.style={draw, font=\sffamily\normalsize\bfseries,minimum size=6mm, scale=0.7, transform shape}]

  \node[main node] (0) {$\emptyset$};
  \node[main node] (1) at (-3,-1) {$\mathtt{f(X_1)}$};
  \node[main node] (2) at (0,-1) {$\mathtt{g(X_1,X_2)}$};
  \node[main node] (3) at (3,-1) {$\mathtt{h(X_1,X_2)}$};
  \node[main node] (4) at (-6,-2) {$\mathtt{f(X_1),f(X_2)}$};
  \node[main node] (5) at (-4.3,-2) {$\mathtt{f(X_1),g(X_2,X_3)}$};
  \node[main node] (6) at (-2.4,-2) {$\mathtt{f(X_1),h(X_2,X_3)}$};
  \node[main node] (7) at (-0.2,-2) {$\mathtt{g(X_1,X_2),g(X_3,X_4)}$};
  \node[main node] (8) at (2.0,-2) {$\mathtt{g(X_1,X_2),h(X_3,X_4)}$};
  \node[main node] (9) at (4.5,-2) {$\mathtt{h(X_1,X_2),h(X_3,X_4)}$};
  
  \path[every node/.style={near start, font=\sffamily\scriptsize,fill=none, inner sep=1pt}]
    (0) edge [] node[] {} (1)
    edge [] node[] {} (2)
    edge [] node[] {} (3)
    (1) edge [] node[] {} (4)
    edge [] node[] {} (5)
    edge [] node[] {} (6)
    (2) edge [] node[] {} (7)
    edge [] node[] {} (8)
    (3) edge [] node[] {} (9);
    
  \end{tikzpicture}
\caption{A lengthening tree of depth 2. $ord(R) = (\mathtt{f/1,g/2,h/2})$.} 
\label{fig:tree-lengthening}
\end{figure}

There are $\binom{d+|R|-1}{d}$ nodes at depth $d$.
Figure~\ref{fig:tree-lengthening} shows a lengthening tree for three relations till depth 2. 

\subsection{Variable Unification}
\label{sec:Variable-Unification}

A property of interest might be expressed by a sentence (conjunction of atoms) with two or more variables referring to the same object.
For instance, the property we should infer might be
\[
\mathsf{P}_{\geq 0.8}\mathsf{F}^{\leq9}\{\mathtt{on(X_1,X_2), cl(X_1), on(X_2,fl)}\}
\]
which involved the property of having a stack of exactly two blocks, or $\mathsf{P}_{\geq 0.8}\mathsf{F}^{\leq 9}\{\mathtt{on(a,X_2),on(X_2,X_4)}\}$ which involved the property of having a stack, with at least two blocks beneath block $\mathtt{a}$.
Consider the latter instance: it would be built up from a node with list $(\mathtt{on/2,on/2})$. 
All combinations of variable names are generated by employing the following variable unification process \textit{before} the object identity framework is enforced.

\begin{figure}
\centering
\begin{tikzpicture}[->,>=stealth',shorten >=1pt,auto,node distance=3cm,
  thick,main node/.style={draw, font=\sffamily\normalsize\bfseries,minimum size=6mm, scale=0.7, transform shape}]

  \node[main node] (0) {$\mathtt{g(X_1,X_2),h(X_3,X_4)}$};
  
  \node[main node] (1) at (-4.5,-1.5) {$\mathtt{g(X_1,X_1),h(X_3,X_4)}$};
  \node[main node] (2) at (-2.5,-2.1) {$\mathtt{g(X_1,X_2),h(X_1,X_4)}$};
  \node[main node] (3) at (-0.8,-1.5) {$\mathtt{g(X_1,X_2),h(X_3,X_1)}$};
  \node[main node] (4) at (0.6,-2.1) {$\mathtt{g(X_1,X_2),h(X_2,X_4)}$};
  \node[main node] (5) at (1.7,-1.5) {$\mathtt{g(X_1,X_2),h(X_3,X_2)}$};
  \node[main node] (6) at (4.5,-2.1) {$\mathtt{g(X_1,X_2),h(X_3,X_3)}$};
  
  \node[main node] (7) at (-5.5,-5.1) {$\mathtt{g(X_1,X_1),h(X_1,X_4)}$};
  \node[main node] (8) at (-4.1,-4.5) {$\mathtt{g(X_1,X_1),h(X_3,X_3)}$};
  \node[main node] (9) at (-2.1,-5.1) {$\mathtt{g(X_1,X_1),h(X_3,X_1)}$};
  \node[main node] (10) at (-1.5,-4.5) {$\mathtt{g(X_1,X_2),h(X_1,X_1)}$};
  \node[main node] (11) at (1.0,-5.1) {$\mathtt{g(X_1,X_2),h(X_1,X_2)}$};
  \node[main node] (12) at (1.8,-4.5) {$\mathtt{g(X_1,X_2),h(X_2,X_1)}$};
  \node[main node] (13) at (4.5,-5.1) {$\mathtt{g(X_1,X_2),h(X_2,X_2)}$};
  
  \node[main node] (14) at (0,-6.8) {$\mathtt{g(X_1,X_1),h(X_1,X_1)}$};
  
  \path[every node/.style={near start, font=\sffamily\scriptsize,fill=none, inner sep=1pt}]
    (0) edge [] node[] {} (1)
    edge [] node[] {} (2)
    edge [] node[] {} (3)
    edge [] node[] {} (4)
    edge [] node[] {} (5)
    edge [] node[] {} (6)
    (1) edge [] node[] {} (7)
    edge [] node[] {} (8)
    edge [] node[] {} (9)
    (2) edge [] node[] {} (10)
    edge [] node[] {} (11)
    edge [] node[] {} (11)
    edge [style={opacity=0.2}] node[] {} (7)
    (3) edge [style={opacity=0.2}] node[] {} (9)
    edge [style={opacity=0.2}] node[] {} (11)
    edge [style={opacity=0.2}] node[] {} (12)
    (4) edge [] node[] {} (12)
    edge [] node[] {} (13)
    edge [style={opacity=0.2}] node[] {} (7)
    (5) edge [style={opacity=0.2}] node[] {} (9)
    edge [style={opacity=0.2}] node[] {} (10)
    edge [style={opacity=0.2}] node[] {} (13)
    (6) edge [style={opacity=0.2}] node[] {} (8)
    edge [style={opacity=0.2}] node[] {} (11)
    edge [style={opacity=0.2}] node[] {} (13)
    (7) edge [] node[] {} (14)
    (8) edge [style={opacity=0.2}] node[] {} (14)
    (9) edge [style={opacity=0.2}] node[] {} (14)
    (10) edge [style={opacity=0.2}] node[] {} (14)
    (11) edge [style={opacity=0.2}] node[] {} (14)
    (12) edge [style={opacity=0.2}] node[] {} (14)
    (13) edge [style={opacity=0.2}] node[] {} (14);
    
  \end{tikzpicture}
\caption{Example of duplicate avoidance for variable unification. 
}
\label{fig:var-unif}
\end{figure}
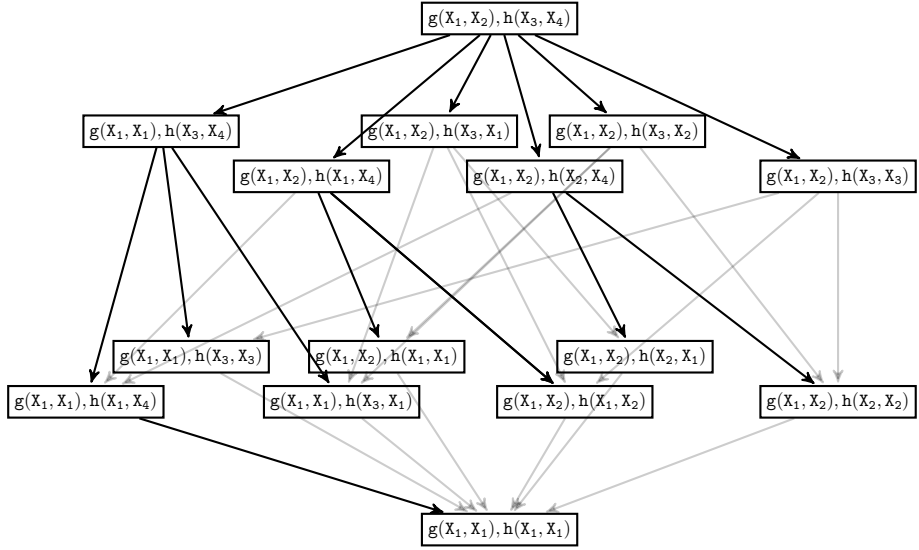

The variable unification process is to take a conjunction with at least two variables with different names, and unify two different variables. Generate new conjunctions by pair-wise unification while possible.
Consider conjunction $\{\mathtt{on(X_1,X_2)}$, $\mathtt{on(X_3,X_4)}\}$ again.
We can unify $\mathtt{X_2}$ and $\mathtt{X_3}$ in two ways, either with substitution $\{\mathtt{X_2}\gets\mathtt{X_3}\}$ or with $\{\mathtt{X_3}\gets\mathtt{X_2}\}$. But the resulting sentences have the same meaning.
In the process of creating variable unifications, we must avoid duplication of meaning.
Taking inspiration directly from \cite{ld04}, ``Of the two variables chosen to be unified, one of them must be not yet unified with any other variables. Moreover, this variable must not be followed by any other already unified variables.''
For instance, Figure~\ref{fig:var-unif} shows a spanning tree of all ways to unify four variables, two at a time.  Following dark arrows is enough to find all unifications. Following light arrows in addition would find the same unifications, causing duplicates.

\subsection{Variable Instantiation}
\label{sec:Variable-Instantiation}


The variable instantiation step comes after the variable unification step. In variable instantiation, a variable in the candidate formula is selected and all variables with the same name are replaced with the same constant. That is, the object identity framework is enforced. Not all variables in a candidate need to be instantiated. It might, for instance, make sense to have a safety property like $\mathsf{P}_{\geq 0.6}\mathsf{F}^{\leq 5}\{\mathtt{on(X,c), cl(X)}\}$, meaning the probability that block $\mathtt{c}$ has a single block on it within five steps is at least 0.6. Or a property like $\mathsf{P}_{\geq 0.6}\mathsf{F}^{\leq 5}\{\mathtt{on(b,X), cl(b)}\}$ also makes sense; the probability that block $\mathtt{b}$ is on some block within five steps, with no block on $\mathtt{b}$ is at least 0.6.

To avoid duplication, we again follow \cite{ld04}: No other arguments to the right of the variable to be instantiated should be the result of a previous instantiation, that is, successive instantiations are performed from left to right.

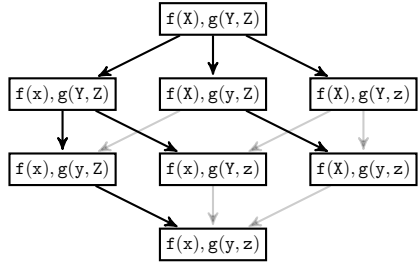
\begin{figure}
\centering
\begin{tikzpicture}[->,>=stealth',shorten >=1pt,auto,node distance=3cm,
  thick,main node/.style={draw, font=\sffamily\normalsize\bfseries,minimum size=6mm, scale=0.7, transform shape}]

  \node[main node] (0) {$\mathtt{f(X),g(Y,Z)}$};
  
  \node[main node] (1) at (-2,-1) {$\mathtt{f(x),g(Y,Z)}$};
  \node[main node] (2) at (0,-1) {$\mathtt{f(X),g(y,Z)}$};
  \node[main node] (3) at (2,-1) {$\mathtt{f(X),g(Y,z)}$};
  
  \node[main node] (4) at (-2,-2) {$\mathtt{f(x),g(y,Z)}$};
  \node[main node] (5) at (0,-2) {$\mathtt{f(x),g(Y,z)}$};
  \node[main node] (6) at (2,-2) {$\mathtt{f(X),g(y,z)}$};
  
  \node[main node] (7) at (0,-3) {$\mathtt{f(x),g(y,z)}$};
  
  \path[every node/.style={near start, font=\sffamily\scriptsize,fill=none, inner sep=1pt}]
    (0) edge [] node[] {} (1)
    edge [] node[] {} (2)
    edge [] node[] {} (3)
    
    (1) edge [] node[] {} (4)
    edge [] node[] {} (5)
    (2) edge [style={opacity=0.2}] node[] {} (4)
    edge [] node[] {} (6)
    (3) edge [style={opacity=0.2}] node[] {} (5)
    edge [style={opacity=0.2}] node[] {} (6)
    (4) edge [] node[] {} (7)
    (5) edge [style={opacity=0.2}] node[] {} (7)
    (6) edge [style={opacity=0.2}] node[] {} (7);
    
  \end{tikzpicture}
\caption{Example of duplicate avoidance for variable instantiation. 
}
\label{fig:var-inst}
\end{figure}

Figure~\ref{fig:var-inst} is an example of the different ways to \textit{uninstantiate} $\mathtt{f(x,y),g(z)}$, where $x,y,z$ represent constants. The result is a spanning tree rooted at $\mathtt{f(x,y),g(z)}$. Light arrows can be pruned; using paths with dark arrows still produce all possible uninstantiations.

But which constants should be used for instantiations? There are actually six spanning trees to consider in the case of $\mathtt{f(X,Y),g(Z)}$ and three constants, and twenty-four spanning trees for four constants. In general, there are $_{|D|}P_v$ permutations and thus so many spanning trees for an expression $C$ for a domain of size $|D|$ and for a sentence with $v=vars(C)$ different variables. For instance, in a domain with three blocks, $a$, $b$ and $c$, instantiating all variables in $\mathsf{F}^{\leq 5}\{\mathtt{on(x,Y), on(Z,v), cl(x)}\}$ in all ways (respecting argument types; where $x$ and $v$ are not variables, but place-holders for constant), results in

\begin{align*}
& \mathsf{F}^{\leq 5}(\mathtt{on(a,Y)\land on(Z,b)\land cl(a)}) \quad \mathsf{F}^{\leq 5}(\mathtt{on(a,Y)\land on(Z,c)\land cl(a)})\\
& \mathsf{F}^{\leq 5}(\mathtt{on(b,Y)\land on(Z,a)\land cl(b)}) \quad \mathsf{F}^{\leq 5}(\mathtt{on(b,Y)\land on(Z,c)\land cl(b)})\\
& \mathsf{F}^{\leq 5}(\mathtt{on(c,Y)\land on(Z,a)\land cl(c)}) \quad \mathsf{F}^{\leq 5}(\mathtt{on(c,Y)\land on(Z,b)\land cl(c)})
\end{align*}

\subsection{Semantic Equivalence of Candidates}
\label{sec:Semantic-Equivalence-of-Candidate}

There is a situation where the single path property might be violated:
Consider candidate\newline $\mathsf{P}_{\geq \alpha}Op^{\leq k}[\mathtt{cl(X),cl(Y)}]$. It will be lengthened to
\[
\mathsf{P}_{\geq \alpha}Op^{\leq k}[\mathtt{cl(X),cl(Y)}, \mathtt{on(X,Z)}] \mbox{ and to } \mathsf{P}_{\geq \alpha}Op^{\leq k}[\mathtt{cl(X),cl(Y),on(Y,Z)}].
\]
But these two formulae are semantically equivalent.


Or consider $\mathsf{P}_{\geq \alpha}Op^{\leq k}\phi$, where $\phi$ is
\[
\phi_1 = \mathtt{[on(X,Y),on(Y,Z),wat(Z)]}
\]
meaning that there is a water container with at least two containers above it. If we are not careful, we might generate a candidate with $\phi$ equals
\[
\phi_2 = \mathtt{[on(X,Y),on(Z,X),wat(Y)]}
\]
with the same meaning as above.
Inspired by \cite{nk03,dr04}, we employ an atom reordering and variable renaming  method to avoid generating candidates with duplicate meaning.

For instance, assume $\mathsf{P}_{\geq \alpha}Op^{\leq k}[\mathtt{cl(X),cl(Y),on(X,Z)}]$ has been generated, and $\mathsf{P}_{\geq \alpha}Op^{\leq k}[\mathtt{cl(X)}$, $\mathtt{cl(Y),on(Y,Z)}]$ is about to be generated. 
With substitution $\{\mathtt{X\gets1,}$ $\mathtt{Y\gets2,}$ $\mathtt{Z\gets3} \}$, $[\mathtt{cl(X),cl(Y)}$, $\mathtt{on(X,Z)}]$ becomes $[\mathtt{cl(1),cl(2),on(1,3)}]$. 
Now by reordering $[\mathtt{cl(X),cl(Y),on(Y,Z)}]$ to $[\mathtt{cl(Y),cl(X)},$ $\mathtt{on(Y,Z)}]$ and applying substitution $\{\mathtt{Y\gets1}, \mathtt{X\gets2}, \mathtt{Z\gets3} \}$, we get $[\mathtt{cl(1),cl(2),on(1,3)}]$ again. Now it is syntactically clear that the two formulae are equivalent.



To apply this check in a principled way, we define a canonical form of a formula.
Given how candidates are generated due to lengthening, the relation symbols appear in a particular order.
Let a formula's \textit{signature} be the concatenation of the formula's relation symbols. For instance, both $[\mathtt{cl(X),cl(Y),on(X,Z)}]$ and $[\mathtt{cl(X),cl(Y),on(Y,Z)}]$ have signature $\mathtt{clclon}$.

Consider all possible reorderings of atoms that maintain its signature. For each such reordering, number variables from left to right.
Take the lexicographically least formula over all possible orderings (and numbered variables) as the canonical form.
For every candidate about to be generated, if numbering its variables from left to right produces that formula's canonical form, then accept it as a candidate, else prune it.


As an illustration of the method, suppose $\phi_1 = \mathtt{[on(X,Y),on(Y,Z),wat(Z)]}$ is being considered as a candidate.
We know that the canonical form of $\phi_1$ is $\mathtt{[on(1,2)}$, $\mathtt{on(2,3),wat(3)]}$ and numbering its variables produces the same formula. $\phi_1$ is thus accepted as a candidate. Because numbering the variables in $\phi_2$ results in $\mathtt{[on(1,2),on(3,1),wat(2)]}$, which is not the canonical form for this signature, $\phi_2$ must not be generated.

\subsection{The Formal Definition}

With the methods for duplicate-avoidance in hand, we are now ready to define our refinement operator $\rho$.


Assuming that $\rho$ conforms to the single-path property, a (search) \textit{tree} of candidate formulae is generated by application of $\rho$ on each node of the tree. Each node represents a candidate $\Psi = \mathsf{P}_{\geq \alpha}\mathit{Op}^{\leq k}\phi$.



Let $chd_\mathit{len}(\phi)$, $chd_\mathit{uni}(\phi)$, $chd_\mathit{ins}(\phi)$ be the children of $\phi$ in the lengthening-tree, applicable unification-tree, respectively, applicable instantiation-tree.
Let $\mathit{last}(\phi)$ be the right-most atom of $\phi$ (added last due to Len).
%

\begin{definition}[Refinement operator for Learn-pCTL]
\label{def:ref-op}
The refinements of any formula $\Psi = \mathsf{P}_{\geq \alpha}\mathit{Op}^{\leq k}\phi$ are produced by applying operations Len, Uni, Ins and Glo to $\Psi$. That is, $\rho(\Psi)$ is the union of
\begin{itemize}
    \item Len: $\{\mathsf{P}_{\geq \alpha}Op^{\leq k}\phi'\mid \phi'\in chd_\mathit{len}(\phi)\} \cup \{\mathsf{P}_{\geq \alpha}Op^{\leq k}\phi''\mid \phi'\in chd_\mathit{len}(\phi)\;\&\;\phi''\in \mathit{Uni}(\phi')\}$, 
    where
    \item $\mathit{Uni}(\phi') = \{\phi''\mid \phi''\in chd_\mathit{uni}(\phi') \;\&$ one of the variables being unified is in $var(\mathit{last}(\phi'))\}$
    \item Ins: $\{\mathsf{P}_{\geq \alpha}Op^{\leq k}\phi'\mid \phi'\in chd_\mathit{ins}(\phi) \;\&$ the variable being instantiated is in $var(\mathit{last}(\phi'))\}$
    \item Glo: $\mathsf{P}_{\geq \alpha}\mathsf{G}^{\leq k}\phi$ if $\mathit{Op}$ is $\mathsf{F}$.
\end{itemize}
\end{definition}

To comply with the {\em single path} property, the first refinement applicable (in the order given above) is applied.
As an example, consider the property $\mathsf{P}_{\geq \alpha}\mathsf{G}^{\leq k}\{\mathtt{on(a,Y)}$, $\mathtt{on(Y,b),cl(a)}\}$. The only way to generate it is by applying Glo to 
$\mathsf{P}_{\geq \alpha}\mathsf{F}^{\leq k}\{\mathtt{on(a,Y)}$, $\mathtt{on(Y,b),cl(a)}\}$. This must be generated by applying Ins to $\mathsf{P}_{\geq \alpha}\mathsf{G}^{\leq k}\{\mathtt{on(a,Y),on(Y,W)}$, $\mathtt{cl(a)}\}$, which must be an Ins application to $\mathsf{P}_{\geq \alpha}\mathsf{G}^{\leq k}\{\mathtt{on(X,Y),on(Y,W),cl(X)}\}$. 
And the latter must come from applying Uni to $\mathsf{P}_{\geq \alpha}\mathsf{G}^{\leq k}\{\mathtt{on(X,Y)}$, $\mathtt{on(Y,W),cl(V)}\}$, which must be a Uni applied to $\mathsf{P}_{\geq \alpha}\mathsf{G}^{\leq k}\{\mathtt{on(X,Y),on(Z,W),cl(V)}\}$. The latter is a realization of a node representing $(\mathtt{on/2,on/2,cl/1})$ in the lengthening-tree, which must be due to a Len of $(\mathtt{on/2,on/2})$ of a Len of $(\mathtt{on/2})$ of a Len of the empty/top formula.
The reader can verify that the only way to obtain $\mathsf{P}_{\geq \alpha}\mathsf{G}^{\leq k}\{\mathtt{on(a,Y),on(Y,b)}$, $\mathtt{cl(a)}\}$ from the empty formula is by applying the refinement operator with the operations in the example in reverse order.
In general, given the restrictions mentioned in the previous sections, there is no opportunity to generate a duplicate formula. That is, for every formula generated, the path from the top formula is unique.

Moreover, the refinement operator complies to the {\em completeness} property:
Each of the four operations makes a minimal refinement to a formula, and every kind of refinement is considered. That is, the only way to specialize a formula is by adding an atom, unifying a variable, instantiating a variable or promoting an $\mathsf{F}$ formula to a $\mathsf{G}$ formula. Every formula is a candidate for further refinement/specialization by one of the four operations, when applicable. Therefore, every formula can be generated, assuming one starts from the top formula (which we do).

Our refinement operator is based closely on the one defined for SeqLogMine \cite{ld04}, and they follow the same reasoning to show that their refinement operator has the two properties for optimality.

\subsection{Domain Knowledge and Argument Types}
\label{sec:Relation-Argument-Types}

As is done in many ILP systems and also by \cite{ld04}, we allow \textit{types} of relation arguments to be specified.
Variables in argument positions of different types may not be unified.
For instance, $\mathtt{X}$ in $\mathtt{on(X,Y)}$ should never be of type \textit{floor}.
And $\mathtt{on(X,X)}$ or $\mathtt{cl(X)\land on(Y,X)}$ does not make sense.

Types can also be used to restrict the constants that are used for the instantiation of variables. Thus, only meaningful values will be used for the substitution.
For instance, for atom $\mathtt{hld(X,Y)}$, $\mathtt{X}$ and $\mathtt{Y}$ must be instantiated with a constant representing a robot, respectively, an object the robot can hold.

Unification and instantiation restrictions due to argument types can significantly reduce the number of candidates that need to be considered.

To summarize, there are at least three opportunities for optimizing the candidate generation process.
(1) One can take advantage of OI-subsumption, (2) one can avoid generating semantically equivalent candidates and (3) one can use domain knowledge to eliminate irrelevant candidates.

\subsection{The Learn-pCTL Algorithm}
\label{sec:policy-agnostic-algo}

Here we present the high-level algorithm of Learn-pCTL. Specifically, this is the algorithm for the policy-agnostic case. The algorithm for the policy-specific case is discussed in Section~\ref{sec:policy-specific-algo}.
\begin{algorithm}
\begin{normalsize}
\caption{The policy-agnostic Learn-pCTL algorithm
\label{alg:check}}
\begin{algorithmic}[1]
\Procedure{Learn-pCTL}{$K, E, \alpha, k, \ell$}
\State Set of solutions $\mathit{Sol}$, initially empty
\For{$\Psi \in$ Len$(\top)$}
    \State Add solutions returned by $\textsc{Search}(K, E, \alpha, k, \ell, \Psi)$ to $\mathit{Sols}$
\EndFor
\Return $\mathit{Sols}$
\EndProcedure
\State
\Procedure{Search}{$K, E, \alpha, k, \ell, \Psi$}
\State Set of solutions $\mathit{Sol}$, initially empty
\If{$\Psi$ is sensible}
    \State $\mathit{Sat_K}(\Psi) \gets$ pCTL-REBEL$(\Psi, K, \alpha, k)$
	\If{$\forall s^+_a\in E^+, \exists s^+_a\in \mathit{Sat_K}(\Psi). s^+_a\preceq s_a$}
        \If{$\forall s^-_a\in E^-, \nexists s_a\in \mathit{Sat_K}(\Psi). s^-_a\preceq s_a$}
    	   \State Add $\Psi$ to $\mathit{Sols}$
    	\EndIf
    	\If{$\mathit{length(\Psi)} < \ell$}
            \For{$\Psi' \in \rho(\Psi)$}
                \State Add solutions returned by $\textsc{Search}(K, E, \alpha, k, \ell, \Psi')$ to $\mathit{Sols}$
            \EndFor
        \EndIf
	\EndIf
\EndIf
\Return $\mathit{Sols}$
\EndProcedure
\end{algorithmic}
\end{normalsize}
\end{algorithm}

The main function that $\textsc{Learn-pCTL}(K, E, \ell)$ calls is $\textsc{Search}(K, E, \ell, \Psi)$, which searches for and returns all formula consistent with examples $E$ of length at most $\ell$. At line 3, Len$(\top)$ refers to the lengthening operation of $\rho$. $K$ is the RMDP modeling the environment/system and is required as input to the pCTL-REBEL model-checker (line 9). At line 8, the algorithm filters out irrelevant candidates. A formula is judged relevant/sensible according to some background or domain knowledge (cf.\ Sec.~\ref{sec:Relation-Argument-Types}). Lines 10 and 11 check consistency of the incumbent candidate with $E$. The indentations below line 10 constitute pruning due to subsumption (cf.\ Sec.~\ref{sec:Pruning-by-Subsumption}). Line 13 stops the algorithm from generating candidates longer than the user-given maximum length $\ell$.



\paragraph{Soundness, Completeness and Termination of Learn-pCTL}
\label{sec:Theoretical-Results}


% ...




We need a couple of definitions before stating our theorems.
\begin{definition}
Example set $E$ is $L^\mathit{FG}$-separable iff there exists a formula $\Psi\in L^\mathit{FG}$ such that for all $s^+\in E^+$, $s^+\models\Psi$ and for all $s^-\in E^-$, $s^-\not\models\Psi$.
\end{definition}
\begin{definition}
$L^\mathit{FG}(m) \doteq \{\Psi\in L^\mathit{FG}\mid length(\Psi)=m\}$.
\end{definition}
\begin{definition}
$Sol^E_d(\alpha,k)$ is the set of all solution formulae (with parameters $\alpha$ and $k$) generated with $d$ refinements and consistent with $E$.
\end{definition}
There is often more than one solution generated with $d$ refinements. 


\begin{theorem}[Soundness]
\label{th:soundness}
Given a set of $L^\mathit{FG}$-separable examples $E$,
if Learn-pCTL finds $\Psi\in Sol^E_d(\alpha,k)$ for any $d>0$, then $\Psi$ is consistent with $E$.
\end{theorem}
\textbf{Proof:}
We have defined consistency of $\Psi\in L^\mathit{FG}$ with $E$ as
\begin{itemize}
    \item $\forall s^+_a\in E^+. \forall s\in I^\Sigma. (s \preceq s^+_a \implies s \models \Psi)$
\item $\forall s^-_a\in E^-. \forall s\in I^\Sigma. (s \preceq s^-_a \implies s \not\models \Psi)$
\end{itemize}
We have implemented consistency for Learn-pCTL as
\begin{itemize}
\item $\forall s^+_a\in E^+, \exists s^+_a\in Sat_K(\Psi). s^+_a\preceq s_a$
\item $\forall s^-_a\in E^-, \nexists s_a\in Sat_K(\Psi). s^-_a\preceq s_a$
\end{itemize}
Proposition~\ref{prp:consistency} proves that the two definitions of consistency are equivalent.
\hfill $\blacksquare$ \vspace{3mm}

\begin{theorem}[Completeness]
\label{th:completeness}
Assuming $\rho$ has the complete property, if $\Psi\in L^\mathit{FG}(m)$ is consistent with $E$ and $m\leq \ell$, where $\ell$ is the user-given maximum solution length, then the Learn-pCTL will find and return $\Psi$.
\end{theorem}
\textbf{Proof:}
Assume $\Psi\in L^\mathit{FG}(m)$ is consistent with $E$ and $m\leq \ell$.
Every formula in $L^\mathit{FG}(m)$ is generated via a series of $d$ refinements employing operator $\rho$.
By the completeness property of $\rho$, $\Psi$ will be generated by Learn-pCTL.
Hence, $\Psi$ will be checked, and will be placed in $Sol^E_d(\alpha,k)$ iff it is consistent with $E$ and $m\leq \ell$ (by Proposition~\ref{prp:consistency}).
\hfill $\blacksquare$ \vspace{3mm}

Theorem~\ref{th:completeness} relies of $\rho$ being complete, which it is.

\begin{theorem}[Termination]
\label{th:termination}
Learn-pCTL will stop and return a (possibly empty) set of solution properties in a finite time.
\end{theorem}
\textbf{Proof:}
There exists a point in the generation and search process of Learn-pCTL, where every candidate to be refined by $\rho$ has length $m = \ell$, where $\ell$ is the user-given maximum solution length. Candidates in $L^\mathit{FG}(\ell)$ cannot be lengthened, and a formula of given length, there are finite unifications and instantiations that can be applied to it. Hence, the Search() procedure will not recurs and Learn-pCTL will return the set of solutions.
\hfill $\blacksquare$ \vspace{3mm}


\section{Experiments}
\label{sec:Experiments}

These experiments are proof-of-concept demonstrations rather than a fully fledged experimental evaluation.
We have a non-optimised implementation that works, that uses pCTL-REBEL, but due to the implementations for both Learn-pCTL and pCTL-REBEL being quite naive and have not been optimised, we do not report on timings.\footnote{The time used by pCTL-REBEL in our framework can be up to nine times longer than the other operations (candidate generation, search and pruning).}


Experiments were run on a 2,3 GHz Dual-Core Intel Core i5 processor, with 16 GB 2133 MHz LPDDR3 memory.

We want to answer the following questions.
\begin{enumerate}
    \item Does Learn-pCTL learn meaningful properties, given a set of examples?
    \item Which and how much pruning occurs?
    \item What is the influence on solution quality for different number of pos./neg. examples?
    \item Does Learn-pCTL learn faster if the refinement operator does not apply instantiation, and are the ungrounded properties learnt still meaningful?
\end{enumerate}

We demonstrate our approach on the Chemical Warehouse (CW) domain.
In the CW domain, we focus on the policy-agnostic case.
Section~\ref{sec:policy-specific-algo} discusses the policy-specific case.

The Blocks World was used to illustrate some concepts earlier in this paper. The Chemical Warehouse (CW) domain is based on the Blocks World where each block is either a water container ($\mathtt{wat/1}$), a rubidium container ($\mathtt{rub/1}$) or a separator ($\mathtt{sep/1}$). Water and Rubidium react explosively with each other and must be kept apart. Hence, if a water container and a rubidium container are in the same stack, they must be separated by a separator.

There are six objects (containers and separators). Objects can be stacked and unstacked (using the $\mathtt{move}$ action). State features are $\mathtt{cl(X)}$, $\mathtt{on(X,Y)}$, $\mathtt{sep(X)}$, $\mathtt{wat(X)}$, $\mathtt{rub(X)}$. The $\mathtt{move}$ action is successful $90\%$ of the time and no effect occurs $10\%$ of the time. However, once a separator has been placed on another object, it cannot be moved again.
The threshold probability is set to $\alpha=0.9$ for all experiments.



For each learning task, we record the number of candidates generated, the number of prunings (due to subsumption, irrelevance and semantic equivalence).

When reporting the properties learned for a given task, we report only the most specific properties, that is, the properties corresponding to the candidates generated with the maximum number of refinements. These are also the most interesting solutions
in concept-learning \cite{mitchell1982generalization}, frequent pattern mining  and  clausal discovery applications \cite{de1997clausal}.

\subsection{Case One}

We generated examples of length eight atoms randomly. That is, each example is generated by random sampling from $\{\mathtt{cl/1,on/2,sep/1,wat/1,rub/1}\}$ and arguments randomly sampled from $\{\mathtt{X_0, \ldots, X_7, a, \ldots, f, fl}\}$ (i.e. eight variables and seven constants, incl.\ the one representing the floor). Only sensible (i.e. physically possible) examples were allowed (cf.\ Sec.~\ref{sec:Relation-Argument-Types}).
For instance, examples containing $\mathtt{on(X,X)}$ or containing $\mathtt{cl(X),on(Y,X)}$ are disallowed.

The target property is
\begin{equation}
    \label{property:1}
    \mathsf{P}_{\geq 0.9}\mathsf{F}^{\leq 3}\mathtt{[on(X0,X1),on(X1,X3),rub(X0),sep(X1),wat(X3)]}.
\end{equation}
That is, an example is considered safe if and only if a Rubidium container is above a water container separated by exactly one separator.
A typical positive example is
\begin{equation}
    \label{ex:1}
    \mathtt{[sep(X1),rub(X6),on(X6,b),wat(g),cl(c),rub(X7),sep(b),on(b,g)]}.
\end{equation}
And a typical negative example is
\begin{equation}
    \label{ex:2}
    \mathtt{[wat(d),cl(X4),wat(e),rub(X3),wat(X6),on(X3,d),on(g,X5),sep(g)]}.
\end{equation}

We ran Learn-pCTL for different amounts of safe and dangerous examples: Four safe examples with zero, four and eight dangerous examples, and eight safe examples with zero, four and eight dangerous examples. That is, we performed experiments for each of six combinations. Table~\ref{tbl:cw-case1-with-inst} reports the statistics.
In all cases, target property \eqref{property:1} was learned (as the most specific solution).


\begin{table*}[t]
\centering
\caption{Results for safety-property 1 (with instantiation).
\label{tbl:cw-case1-with-inst}
}
\footnotesize
\begin{tabular}{| p{35pt} || p{40pt}  p{40pt}  p{40pt}  p{40pt} |}
\hline
 &  & \multicolumn{3}{c|}{Pruning} \\
\cline{3-5}
$|E^+|/|E^-|$ & \# Cands. & Subsump. & Irrel. & Sem. Eqv. \\
\hline\hline
4/0 & 316($\pm$ 83) & 282($\pm$ 65) & 33($\pm$ 0) & 25($\pm$ 22)  \\
4/4 & 386($\pm$ 64) & 336($\pm$ 55) & 44($\pm$ 6) & 15($\pm$ 8)  \\
4/8 & 268($\pm$ 11) & 235($\pm$ 10) & 41($\pm$ 7) & 12($\pm$ 2) \\
8/0 & 255($\pm$ 4) & 222($\pm$ 3) & 33($\pm$ 0) & 10($\pm$ 0) \\
8/4 & 262($\pm$ 22) & 229($\pm$ 21) & 41($\pm$ 8) & 10($\pm$ 0)  \\
8/8 & 253($\pm$ 5) & 220($\pm$ 4) & 40($\pm$ 7) & 10($\pm$ 0)  \\
\hline 
\end{tabular}
\end{table*}

Next, we perform a set of experiments on the same examples generated before, the only difference is that now, Learn-pCTL does not apply instantiation. The idea is that when several constants are involved in the same general property, then the property learned will be general enough to subsume all the more specific (more grounded) examples. For instance, given two positive examples $[\mathtt{cl(X),on(X,a)}]$ and  $[\mathtt{cl(X),on(X,b)}]$, the most specific property consistent with them is $[\mathtt{cl(X),on(X,Y)}]$. But if the user expects or requires only non-grounded properties, then no candidate mentioning a constant needs to be generated. Table~\ref{tbl:cw-case1-without-inst} reports the statistics.
As expected, the properties learned are identical to those learned in the case where instantiation was applied.
And we see that Learn-pCTL is always slightly faster than with instantiation.

\begin{table*}[t]
\centering
\caption{Results for safety-property 1 \textit{without} instantiation.
\label{tbl:cw-case1-without-inst}
}
\footnotesize
\begin{tabular}{| p{35pt} || p{40pt}  p{40pt}  p{40pt}  p{40pt}|}
\hline
 &  & \multicolumn{3}{c|}{Pruning} \\
\cline{3-5}
$|E^+|/|E^-|$ & \# Cands. & Subsump. & Irrel. & Sem. Eqv. \\
\hline\hline
4/0 & 214($\pm$ 56) & 171($\pm$ 43) & 33($\pm$ 0) & 22($\pm$ 17)  \\
4/4 & 245($\pm$ 39) & 194($\pm$ 30) & 41($\pm$ 8) & 15($\pm$ 8)   \\
4/8 & 175($\pm$ 8) & 141($\pm$ 7) & 33($\pm$ 0) & 12($\pm$ 2)   \\
8/0 & 166($\pm$ 2) & 133($\pm$ 1) & 33($\pm$ 0) & 16($\pm$ 9) \\
8/4 & 170($\pm$ 11) & 137($\pm$ 9) & 33($\pm$ 0) & 10($\pm$ 0)  \\
8/8 & 165($\pm$ 2) & 133($\pm$ 2) & 33($\pm$ 0) & 10($\pm$ 0)  \\
\hline 
\end{tabular}
\end{table*}



\subsection{Case Two}

The setup in Case Two situation is very similar to that of Case One, except for the definition of the safety property: In Case Two, an example is considered safe if two Rubidium containers are stacked directly on each other, or if two water containers are stacked directly on each other (touching). 
In other words, the target property is
\begin{equation}
    \label{property:2}
    \mathsf{P}_{\geq 0.9}\mathsf{F}^{\leq 3}\mathtt{[on(X0,X1),rub(X0),rub(X1)]\lor[on(X2,X3),wat(X2),wat(X3)]}.
\end{equation}

This property cannot be described by (state) formulae in $L^{FG}$ due to the need for a notion of disjunction. Nonetheless, we found that informative properties are learned when no positive example mentions both disjuncts.

Example~\ref{ex:3} is such an instance.
\begin{equation}
    \label{ex:3}
    \mathtt{[cl(X2),cl(e),cl(X2),on(X6,d),on(X2,X3),rub(X3),cl(X7),rub(X2)]}.
\end{equation}
Table~\ref{tbl:cw-case2-with-inst} reports the statistics for the standard Learn-pCTL (with instantiation).
For $|E^+|/|E^-| = 4/0$, for each of the four tasks, one property was learned. They are, respectively,
\begin{align*}
    & \mathsf{P}_{\geq 0.9}\mathsf{F}^{\leq 3}\mathtt{[on(X0,X1)]}\\
    & \mathsf{P}_{\geq 0.9}\mathsf{F}^{\leq 3}\mathtt{[cl(X0),wat(X1)]}\\
    & \mathsf{P}_{\geq 0.9}\mathsf{F}^{\leq 3}\mathtt{[on(X0,X1),rub(X2),rub(X3)]}\\
    & \mathsf{P}_{\geq 0.9}\mathsf{F}^{\leq 3}\mathtt{[cl(X0),on(X1,X2),rub(X2),rub(X1),rub(X5)]}
\end{align*}
The properties range between uninformative (top formula) and reasonably informative (bottom formula). The uninformative cases occur when there exists an example in $E^+$ with two Rubidium containers touching and another example in $E^+$ with two water containers touching. $\mathtt{[on(X0,X1)]}$ and $\mathtt{[cl(X0),wat(X1)]}$ are the `common' parts of the examples in the respective uninformative cases. In the informative cases, no example mentioned two water containers touching.

For $|E^+|/|E^-| = 8/0$, for three of the four tasks, one property was learned and for the other task, two properties were learned. They are all uninformative. This is because with eight positive examples, there is a higher likelihood that there is an example where two Rubidium containers touch \textit{and} an example where two water containers touch.
For $|E^+|/|E^-| = 8/4$ and $|E^+|/|E^-| = 8/8$, there were no properties consistent with $E$. In these cases, even the `common' parts of the positive examples are filtered out by the negative examples.

\begin{table*}[t]
\centering
\caption{Results for safety-property 2 (with instantiation).
\label{tbl:cw-case2-with-inst}
}
\footnotesize
\begin{tabular}{| p{35pt} || p{40pt}  p{40pt}  p{40pt}  p{40pt} |}
\hline
 &  & \multicolumn{3}{c|}{Pruning} \\
\cline{3-5}
$|E^+|/|E^-|$ & \# Cands. & Subsump. & Irrel. & Sem. Eqv. \\
\hline\hline
4/0 & 124($\pm$ 93) & 112($\pm$ 82) & 11($\pm$ 5) & 28($\pm$ 37)  \\
4/4 & 61($\pm$ 33) & 54($\pm$ 25) & 11($\pm$ 5) & 0.5($\pm$ .75) \\
4/8 & 109($\pm$ 31) & 101($\pm$ 29) & 36($\pm$ 20) & 7($\pm$ 4)  \\
8/0 & 52($\pm$ 18) & 49($\pm$ 17) & 11($\pm$ 6) & 0($\pm$ 0)  \\
8/4 & 33($\pm$ 4) & 31($\pm$ 4) & 8($\pm$ 1) & 0($\pm$ 0)  \\
8/8 & 35($\pm$ 5) & 33($\pm$ 4) & 9($\pm$ 1) & 0($\pm$ 0)  \\
\hline 
\end{tabular}
\end{table*}

Table~\ref{tbl:cw-case2-without-inst} reports the statistics for Learn-pCTL without instantiation applied, using the same examples as used for the experiments with instantiation. The properties learned are exactly the same as for with instantiation.

\begin{table*}[t]
\centering
\caption{Results for safety-property 2 \textit{without} instantiation.
\label{tbl:cw-case2-without-inst}
}
\footnotesize
\begin{tabular}{| p{35pt} || p{40pt}  p{40pt}  p{40pt}  p{40pt} |}
\hline
 &  & \multicolumn{3}{c|}{Pruning}\\
\cline{3-5}
$|E^+|/|E^-|$ & \# Cands. & Subsump. & Irrel. & Sem. Eqv.\\
\hline\hline
4/0 & 97($\pm$ 84) & 84($\pm$ 73) & 12($\pm$ 5) & 28($\pm$ 37) \\
4/4 & 35($\pm$ 19) & 28($\pm$ 12) & 11($\pm$ 6) & 0.5($\pm$ .75)  \\
4/8 & 65($\pm$ 18) & 58($\pm$ 16) & 31($\pm$ 19) & 7($\pm$ 4) \\
8/0 & 26($\pm$ 10) & 23($\pm$ 9) & 11($\pm$ 6) & 0($\pm$ 0)  \\
8/4 & 18($\pm$ 2) & 17($\pm$ 2) & 7($\pm$ 0) & 0($\pm$ 0)  \\
8/8 & 19($\pm$ 2) & 17($\pm$ 2) & 7($\pm$ 0) & 0($\pm$ 0)  \\
\hline 
\end{tabular}
\end{table*}

\subsection{Case Three}

Notice that all the properties learned in the cases above are Eventually formulae. Here we want to confirm that Globally formulae can be learned. To do this, we start with a property of the form $\Psi = \mathsf{P}_{\geq \alpha}\mathsf{G}^{\leq k}\phi$, use pCTL-REBEL to find $Sat_K(\Psi)$, then set $E^+ = Sat_K(\Psi)$ and $E^- = \emptyset$.

We ran Learn-pCTL without instantiation on a six-atom, randomly generated state-formula based on the safety property from Case One:
\[\Psi_{rnd} = \mathsf{P}_{\leq0.9} \mathsf{G}^{\leq 3}[\mathtt{on(X3,c),sep(X3),on(b,X3),wat(c),on(e,X4),rub(b)}].\]
and on a manually chosen formula:
\[\Psi_{mnl} = \mathsf{P}_{\leq0.9} \mathsf{G}^{\leq 3}[\mathtt{on(W,X),on(X,R), wat(W), sep(X), rub(R)}].\]

%
\begin{align*}
Sat_K(\Psi_{rnd}) &= \{[\mathtt{cl(e),on(e,fl),on(b,S),on(S,c),rub(b),sep(e),sep(S),wat(c)}],\\
& [\mathtt{cl(e),cl(X),on(b,S),on(e,Y),on(S,c),rub(b),sep(S),wat(c),wat(e)}],\\
& [\mathtt{cl(e),cl(X),on(b,S),on(e,Y),on(S,c),rub(b),rub(e),sep(S),wat(c)}]\}
\end{align*}
\begin{align*}
Sat_K(\Psi_{mnl}) &= \{[\mathtt{cl(Y),on(Y,fl),on(X,R),on(W,X),rub(R),sep(Y),sep(X),wat(W)}],\\
& [\mathtt{cl(W1),cl(Z),on(W1,Y),on(X,R),on(W2,X),rub(R),sep(X),wat(W1),wat(W2)}],\\
& [\mathtt{cl(R1),cl(Z),on(R1,Y),on(X,R2),on(W,X),rub(R1),rub(R2),sep(X),wat(W)}]\}
\end{align*}

\begin{table*}[t]
\centering
\caption{Results for learning target property $\Psi_{rnd}$.
\label{tbl:cw-case3.1}
}
\small
\begin{tabular}{| p{40pt} || p{30pt}  p{40pt}  p{40pt}  p{40pt} |}
\hline
 &  & \multicolumn{3}{c|}{Pruning}\\
\cline{3-5}
$\Psi$ & \# Cands. & Subsump. & Irrel. & Sem. Eqv.\\
\hline\hline
$\Psi_{rnd}$ & 1041 & 893 & 426 & 191  \\
$\Psi_{mnl}$ & 1083 & 935 & 426 & 219  \\
\hline 
\end{tabular}
\end{table*}
Table~\ref{tbl:cw-case3.1} reports the statistics. The properties learned for targets $\Psi_{rnd}$ and $\Psi_{mnl}$ were, respectively,
\begin{align*}
    &\mathsf{P}_{\geq 0.9}\mathsf{G}^{\leq 3}\mathtt{[cl(X0),on(X1,X2),on(X0,X4),on(X2,X6),rub(X1),sep(X2),wat(X6)]},\\
    &\mathsf{P}_{\geq 0.9}\mathsf{G}^{\leq 3}\mathtt{[cl(X0),on(X1,X2),on(X0,X4),on(X2,X6),rub(X6),sep(X2),wat(X1)]},
\end{align*}
which are subsumed by their targets.

pCTL-REBEL can take several tens of seconds to compute $Sat_K(\Psi)$ when $\Psi$ is a Globally formula, which, by design, occurs more often in this experiment. pCTL-REBEL has only recently been proposed and has not yet been optimized for speed.

\subsection{Answering the Experimental Questions}

\begin{enumerate}
    \item Q: Does Learn-pCTL learn meaningful properties, given a set of examples? A: Yes (when examples are properly labeled).
    \item Q: Which and how much pruning occurs? A: A significant amount of pruning occurs, especially due to subsumption. The algorithm would be infeasible with out pruning.
    \item Q: What is the influence on solution quality for different number of pos./neg. examples? A: We see that as the number of positive examples increases, the number of candidates generated decreases (and thus the running time decreases). There seems to be no correlation between number of negative examples and candidates generated / running time. This is not surprising, since subsumption pruning is based on positive examples.
   \item Q: Does Learn-pCTL learn faster if the refinement operator does not apply instantiation, and are the ungrounded properties learnt still meaningful? Yes. We observed that when candidates are not instantiated, Learn-pCTL is always faster than with instantiation. However, the speed-up is typically not significant.
\end{enumerate}

\section{Using Learn-pCTL with a Known Policy}
\label{sec:policy-specific-algo}


We have introduced Learn-pCTL, an algorithm that learns a set $Sol^E_d(\alpha,k)$ of relational pCTL formulae over given positive states $E^+$ and negative states $E^-$. 
Learn-pCTL has two settings: policy-agnostic and policy-specific. The policy-specific setting is related to formulae of the form $\mathsf{P}^\pi_{\geq \alpha}\Phi$ which is defined in Section~\ref{sec:Semantics-of-pCTL-REBEL}.
Policy-agnostic Learn-pCTL (\textit{a-Learn-pCTL} for short) had more of the focus until here.
However, we argue that some situations require a policy-specific version of Learn-pCTL. We do not give experiments, but we discuss a potential use case and give a naive algorithm by slightly modifying a-Learn-pCTL. 

\paragraph{Why policy-specific Learn-pCTL?}

Recall that a-Learn-pCTL focuses on learning properties that are human readable. This setting is useful when the domain expert gives labeled states (examples), and then selects a property that best suits the domain, without having a specific policy in mind. a-Learn-pCTL returns formulae that may have underlying, implicit policies that are distinct from one another. 

However, when the policy is known, using a-Learn-pCTL is not ideal since it may induce formulae that can never be satisfied by the known policy.
A policy-specific setting is required to learn pCTL formulae with respect to a policy. 
For example, the domain expert may want to know the expected revenue target within the next month by following \textit{the marketing policy at hand}. 
Policy-specific Learn-pCTL (\textit{s-Learn-pCTL} for short) is similar to a-Learn-pCTL, except that it must take as an input a policy to prune inconsistent formulae. Essentially, s-Learn-pCTL returns a subset of the formulae that would have been returned by a-Learn-pCTL. 

\paragraph{Can Learn-pCTL learn useful, human-readable properties of a particular policy?}
Turning a-Learn-pCTL into s-Learn-pCTL can be naively done by slightly modifying the consistency checking step (cf.\ Sec.~\ref{sec:Consistency-Checking}), specifically, the input to pCTL-REBEL. Instead of giving as an input a RMDP as in Algorithm~\ref{alg:check}, we first convert the RMDP into a policy-specific RMDP that allows for one and only one action in each state. 
A policy-specific RMDP is analogous to a Markov Chain. 
Notice, this conversion requires the given policy to be deterministic and memoryless. 
\textsc{s-Learn-pCTL} (Algo.~\ref{alg:check2}) is the new algorithm. The blue text highlights where the difference is. In particular, s-Learn-pCTL and Search now also take a policy $\pi$ as input. And at line 9, constrainRMDP() returns an RMDP model constrained by $\pi$, which is input to the model-checker at line 10.

An optimization opportunity to this naive policy-specific algorithm lies in the candidate generation step: If one could generate only candidates consistent with the given policy, then the search process would be significantly more efficient. But we leave this for the future.

\begin{algorithm}
\begin{normalsize}
\caption{The \textcolor{blue}{policy-specific} Learn-pCTL algorithm
\label{alg:check2}}
\begin{algorithmic}[1]
\Procedure{s-Learn-pCTL}{$K, E, \alpha, k, \ell$, \textcolor{blue}{$\pi$}}
\State Set of solutions $\mathit{Sol}$, initially empty
\For{$\Psi \in$ Len$(\top)$}
    \State Add solutions returned by $\textsc{Search}(K, E, \alpha, k, \ell, \Psi$, \textcolor{blue}{$\pi$}) to $\mathit{Sols}$
\EndFor
\Return $\mathit{Sols}$
\EndProcedure
\State
\Procedure{Search}{$K, E, \alpha, k,, \ell, \Psi$, \textcolor{blue}{$\pi$}}
\State Set of solutions $\mathit{Sol}$, initially empty
\If{$\Psi$ is sensible}
    \State \textcolor{blue}{$K'\gets$  constrainedRMDP($\pi$, $K$, $\alpha, k$)}\;
    \State $\mathit{Sat_K}(\Psi) \gets$ pCTL-REBEL$(\Psi,$ \textcolor{blue}{$K'$}, $\alpha, k$)\;
	\If{$\forall s^+_a\in \mathcal{T}^+, \exists s^+_a\in \mathit{Sat_K}(\Psi). s^+_a\preceq s_a$}
        \If{$\forall s^-_a\in \mathcal{T}^-, \nexists s_a\in \mathit{Sat_K}(\Psi). s^-_a\preceq s_a$}
    	   \State Add $\Psi$ to $\mathit{Sols}$
    	\EndIf
         \If{$\mathit{length(\Psi)} < \ell$}
            \For{$\Psi' \in \rho(\Psi)$}
                \State Add solutions returned by $\textsc{Search}(K, E, \alpha, k, \ell, \Psi', \textcolor{blue}{\pi})$ to $\mathit{Sols}$
            \EndFor
        \EndIf
	\EndIf
\EndIf
\Return $\mathit{Sols}$
\EndProcedure
\end{algorithmic}
\end{normalsize}
\end{algorithm}

\paragraph{What kind of insight does Learn-pCTL give to revise or give advice about the underlying policy?}

s-Learn-pCTL takes a set of labelled states and a (deterministic, memoryless) policy, and returns pCTL formulae that are consistent to the policy and classify the states. It can be used to extract (human-interpretable) properties of a (non-human-interpretable) policy. 
A use case of s-Learn-pCTL is to help the domain expert to tune the policy. A domain expert can iteratively modify the policy, activate Learn-pCTL, and observe the changes in the learned formulae. 

\medskip
To illustrate the use case further, consider the following scenario. In an automated warehouse, given a set of good scenarios (e.g. picking up a box, delivering a box) and bad scenarios (e.g. crashing into a wall or another robot). Given an initial policy that an autonomous robot tries to turn left when another robot is ahead in the aisle. s-Learn-pCTL returns the property of the policy that the autonomous robot does not crash within the next 10 minutes with a probability of 0.5 (i.e. $\mathsf{P}^\pi_{\geq 0.5}G^{<10} no\_crash$). However, given this property, the domain expert recognizes the policy is not safe enough, thus modifies the policy so that a robot should instead try to turn right when detecting another robot ahead. Then, s-Learn-pCTL returns a property of the modified policy that the robot will stay safe within the next 10 minutes with a higher probability of 0.7 (i.e. $\mathsf{P}^\pi_{\geq 0.7}G^{<10} no\_crash$).

\section{Related Work}

There are several publications on inferring or mining temporal properties from examples. The properties are typically expressed in the syntax of a useful temporal logic.
Some approaches learn from only positive examples and others from both positive and negative examples. Some languages of interest involve continuous parameters, but seldom probabilities. 
When it comes to inference/mining of temporal properties, the literature discusses two families of logics, broadly speaking: linear temporal logic (LTL) \cite{lpb15,ks17,ng18,r19,cm19,xojt19} and signal temporal logic (STL) \cite{kjagb14,bbs14,bvpyb16,vidbwdb17,bdmj20}.

LTL is a \textit{linear-time} logics, that is, its formulae are interpreted in terms of sequences of states (paths). 
Signal temporal logic (STL)
is a temporal logic defined over \textit{signals} (continuous-time, continuous-valued functions from $\mathbb{R}^+$ to $\mathbb{R}^n$) and their trajectories.
This is in contrast to CTL formulae, which are interpreted in terms of a state $s$ and the possible paths that can occur from it. That is, CTL formulae are interpreted in terms of the \textit{computation tree} rooted at a state $s$.

\cite{wz09} combined static analysis with model checking to mine CTL formulae from program code to describe operational preconditions in programs. They do neither consider probabilities nor relational representations, and their setting does not assume a set of examples to be given.

All these related works learn properties (i.e. formulae) that match a given set or family of template formulae, as does ours. The fragment matching the templates is supposed to be useful for a particular kind of domain. However, none of the related works are concerned with \textit{probabilistic} temporal logics, whereas ours includes probabilities, and none involve a relational language; they are all propositional.

SeqLog is a logical language for mining and querying sequential data and databases \cite{ld04}. The elements of a sequence are logical, relational atoms. MineSeqLog is an Inductive Logic Programming algorithm for mining for SeqLog patterns of interest in sequential data. It combines principles of pattern mining with an optimal \textit{refinement operator}. As mentioned in the introduction, our refinement operator is adapted from the one defined for MineSeqLog.
And \cite{dfgl10} define ``multi-sequential patterns which are first-order temporal patterns'' and two algorithms for mining frequent patterns.

There is the topic of \textit{relational sequence learning} \cite{kdgkl08}, which includes tasks such as sequential pattern mining and sequence classification.
SeqLog \cite{ld04} and the work of \cite{dfgl10} are related to the former task.
%
The main differences of MineSeqLog and the work of \cite{dfgl10} to Learn-pCTL are that (1) they operates on sequences, whereas Learn-pCTL operates on states (the start state of a sequence), (2) there is a probabilistic element in pCTL, but not in the related works and (3) they search for properties describing frequent patterns, whereas we search for properties describing reachable states.

Related to the latter task, Lynx is a relational pattern-based \textit{classifier} \cite{dbfe11}. They employ probabilistic models for relational sequence learning to predict a category for an (unlabeled) sequence, but probabilities are not mentioned in the sequence language itself.
Moreover, classification is even farther away from the task we tackle in this work than is pattern mining.

\section{Conclusion}

We presented a system called Learn-pCTL which learns non-nested Eventually and Globally pCTL formulae (properties), given a set of abstract relational states labeled as safe or dangerous. The states in question are assumed to be from a relational Markov decision process.
Our contribution is the first method for inductively learning a relational probabilistic CTL formula from a set of positive and negative examples in an RMDP setting.
It works in two modes, one where the policy is fixed and the other where any policy can be used. 



We have implemented Learn-pCTL and demonstrated that it works as expected, in the policy-agnostic mode. We also discussed how the algorithm could be used in the policy-specific mode.

There are several ways in which this work can be extended. One could allow for nested formulae, e.g., with nested probability operators.
One could infer step-bounds and threshold probabilities instead of fixing them.

Having no disjunction in state formulae severely limits what Learn-pCTL can learn.
Adding disjunction is conceptually simple and would make Learn-pCTL significanly more expressive. It is thus the first extension one should consider for this work.

Finally, 
pCTL-REBEL has been developed as an academic study and has not been optimized for practical use. Therefore, if pCTL-REBEL could be made an order of magnitude faster (which seems possible), then Learn-pCTL would also become an order of magnitude faster - Learn-pCTL calls pCTL-REBEL several tens to several hundred times per learning task. The generation and search processes of Learn-pCTL can also still be optimized. However, these are mostly software engineering issues.



%
%

\printbibliography

\end{document}